\documentclass[times,twocolumn,final]{elsarticle}

\usepackage{medima_arxiv}
\usepackage{framed,multirow}

\usepackage{amssymb}
\usepackage{latexsym}

\usepackage{url}
\usepackage{xcolor}
\usepackage{moresize}
\usepackage{booktabs}
\usepackage{colortbl}
\usepackage{xcolor}
\usepackage{graphicx}
\usepackage{multirow}
\usepackage{subfigure}
\usepackage{amsmath}
\usepackage[colorlinks=true,citecolor=newtext,urlcolor=newtext]{hyperref}

\usepackage{pifont}
\newcommand{\xmark}{\ding{55}}%

\definecolor{tab}{rgb}{0.9,0.9,0.97}
\newcolumntype{g}{ >{\columncolor{tab}} c }
\definecolor{newcolor}{rgb}{.8,.349,.1}
\definecolor{dgr}{RGB}{20, 110, 50}
\definecolor{newtext1}{RGB}{0, 0, 0}
\definecolor{changetext1}{RGB}{0, 0, 0}
\definecolor{newtext}{RGB}{0, 0, 0}
\definecolor{changetext}{RGB}{0, 0, 0}
\journal{Medical Image Analysis}

\begin{document}

\verso{This article has been accepted for publication in Medical Image Analysis. DOI: 10.1016/j.media.2023.102844}
\begin{frontmatter}

\title{Dissecting Self-Supervised Learning Methods for Surgical Computer Vision}

\author[1,3]{Sanat \snm{Ramesh}\fnref{fn1}}
\author[1]{Vinkle \snm{Srivastav}\corref{cor1}\fnref{fn1}}
\cortext[cor1]{Corresponding author: 
  Tel.: +33-039-041-3553;}
\ead{srivastav@unistra.fr}
\fntext[fn1]{Sanat Ramesh, Vinkle Srivastav, Deepak Alapatt and Tong Yu contributed equally and share co-first authorship.}
\author[1]{Deepak \snm{Alapatt}\fnref{fn1}}
\author[1]{Tong \snm{Yu}\fnref{fn1}}
\author[1]{Aditya \snm{Murali}}
\author[1,5]{Luca \snm{Sestini}}
\author[1]{Chinedu Innocent \snm{Nwoye}}
\author[1]{Idris \snm{Hamoud}}
\author[1]{Saurav \snm{Sharma}}
\author[2]{Antoine \snm{Fleurentin}}
\author[1,2]{Georgios \snm{Exarchakis}}
\author[1,2]{Alexandros \snm{Karargyris}}
\author[1,2]{Nicolas \snm{Padoy}}

\address[1]{ICube, University of Strasbourg, CNRS, Strasbourg 67000, France}
\address[2]{IHU Strasbourg, Strasbourg 67000, France}
\address[3]{Altair Robotics Lab, Department of Computer Science, University of Verona, Verona 37134, Italy}
\address[5]{Department of Electronics, Information and Bioengineering, Politecnico di Milano, Milano 20133, Italy}

\received{1 May 2013}
\finalform{10 May 2013}
\accepted{13 May 2013}
\availableonline{15 May 2013}
\communicated{S. Sarkar}

\begin{abstract}
The field of surgical computer vision has undergone considerable breakthroughs in recent years with the rising popularity of deep neural network-based methods. However, standard fully-supervised approaches for training such models require vast amounts of annotated data, imposing a prohibitively high cost; especially in the clinical domain. Self-Supervised Learning (SSL) methods, which have begun to gain traction in the general computer vision community, represent a potential solution to these annotation costs, allowing to learn useful representations from only unlabeled data. Still, the effectiveness of SSL methods in more complex and impactful domains, such as medicine and surgery, remains limited and unexplored.
In this work, we address this critical need by investigating four state-of-the-art SSL methods (MoCo v2, SimCLR, DINO, SwAV) in the context of surgical computer vision. We present an extensive analysis of the performance of these methods on the Cholec80 dataset for two fundamental and popular tasks in surgical context understanding, phase recognition and tool presence detection. We examine their parameterization, then their behavior with respect to training data quantities in semi-supervised settings. Correct transfer of these methods to surgery, as described and conducted in this work, leads to substantial performance gains over generic uses of SSL - up to {\color{changetext}7.4\%} on phase recognition and 20\% on tool presence detection - as well as state-of-the-art semi-supervised phase recognition approaches by up to 14\%. {\color{newtext} Further results obtained on a highly diverse selection of surgical datasets exhibit strong generalization properties.} The code is available at \url{https://github.com/CAMMA-public/SelfSupSurg}.
\end{abstract}


\end{frontmatter}


\section{Introduction}

\begin{figure*}[t]
\centerline{\includegraphics[width=0.8\linewidth]{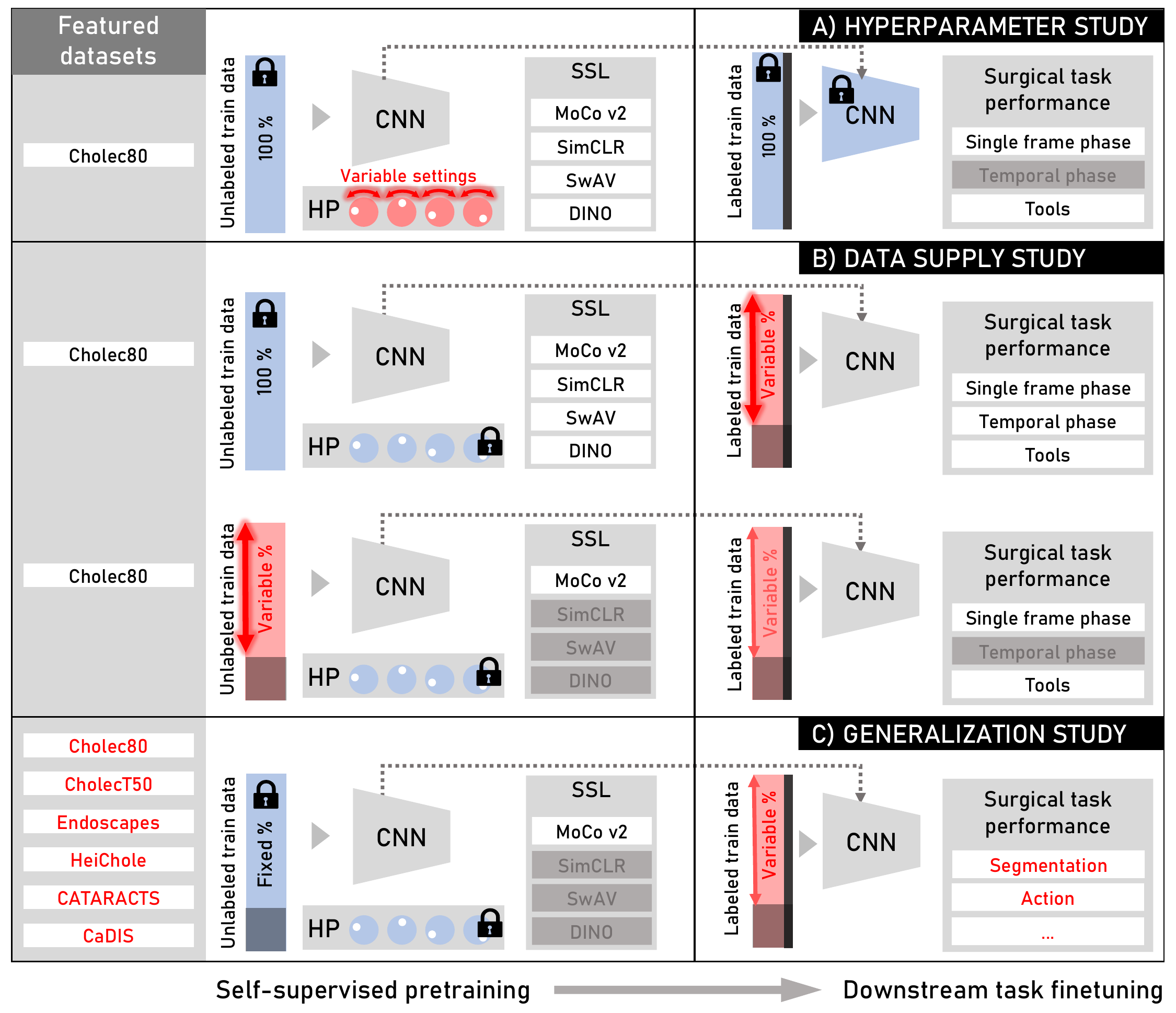}}
\caption{{\color{changetext} Three} stages of the study: (A) Hyperparameter study: Analyzing the influence of hyperparameters when adapting SSL methods to the surgical domain. (B) Data supply study: Evaluating the response of SSL methods to varying amounts of (1) labeled and (2) unlabeled data. {\color{newtext} (C) Generalization study: observing how well SSL generalizes to a much larger variety of surgical data and tasks.}}
\label{fig:concept}
\end{figure*}

Automatic analysis and interpretation of visual signals from the operating room (OR) is the primary concern of surgical computer vision, a fast-growing discipline that is expected to play a major role in the development of reliable decision support systems for surgeons \citep{sds}. Recent developments in the field have indeed resulted in increasingly refined vision algorithms; however, a majority of these studies have only been conducted on datasets containing small amounts of recorded procedures, all of which have been manually annotated by clinical experts. In future developments, much larger quantities of data will be required in order to account for variations in anatomy, patient demographics, clinical workflow, surgical skills, instrumentation, and image acquisition \citep{sds_2}.

For that purpose, raw video data can be supplied on a very large scale by laparoscopic surgeries, since they are guided by intra-abdominal video streams: in the United States, nearly 1M laparoscopic cholecystectomies are performed each year, resulting in approximately 630k hours of footage for just this one type of procedure. Yet, datasets used for training current surgical vision models remain disproportionately small. For example, Cholec80 \citep{twinanda2016endonet}, one of the most popular datasets in the field \citep{sds}, hardly exceeds 50 hours of recordings. Apart from medico-legal constraints, the critical factor leading to this sparsity of data is the reliance on manual annotations. While labels for natural images can be easily supplied by the general public, surgical annotations usually require clinical expertise. As a result, the fully supervised approach - i.e. training models with entirely annotated datasets - may prove to be unsustainable in surgical computer vision.

In computer vision, an alternative has emerged in the form of Self-Supervised Learning (SSL) \citep{ssl_survey}. Considerable progress has been made in this area, with increasingly refined methods for extracting rich vector representations from images without labels, using only the raw pixel data. This research topic has so far not been thoroughly explored in surgical applications. In the few self-supervised training tasks proposed by the community, learning from the visual content itself is generally de-emphasized in favor of utilizing other available sources of information - for example time \citep{funke2018temporal, yengera2018less}, stereoscopy \citep{yang2021real} or robot kinematics \citep{sestini2021kinematic}. State-of-the-art natural image SSL methods, with their advanced representational capabilities, have yet to be adequately demonstrated on surgical images.

However expanding SSL methods outside of natural images can be challenging, especially in a complex domain such as surgery. Most notably, heavy parameter tuning based on heuristics \citep{xiao2020should} might be required. Robustness against large variations in domains and tasks also is not guaranteed; in-depth performance analysis has essentially been conducted on general computer vision datasets \citep{feichtenhofer2021large}, most commonly Imagenet, which contains 14M images and over 1000 visually distinct classes. In contrast, Cholec80, one of the most prominent surgical computer vision datasets \citep{sds}, contains 80 videos of procedures resulting in under 200k frames at 1fps. Only 7 classes of surgical phases and 7 classes of tools are featured; moreover, the visual evidence to distinguish them is highly sparse, especially for time-based tasks such as surgical phase recognition, a coarse-grained form of activity recognition. Further, since surgical videos can last up to several hours depicting a relatively stable scene, it is non-trivial to determine how existing SSL frameworks can best accommodate frames coming from the same procedure. Finally, these issues may be exacerbated by surgery-specific confounding factors such as smoke, bleeding, occlusions, or rapid tool movements. Such fundamental differences between natural and surgical image data motivate the need for a thorough study of SSL in the surgical domain.

The work presented here thoroughly addresses this need in {\color{changetext} three} distinct steps (see Fig. \ref{fig:concept}). We select four SSL methods - MoCo v2 \citep{chen2020improved}, SimCLR \citep{chen2020simple}, SwAV \citep{caron2020unsupervised}, DINO \citep{caron2021emerging} - suitably covering the state of the art in general computer vision, and extensively examine hyperparameter variations for each of them on Cholec80. We identify key differences with the natural image domain, highlighting hyperparameter tuning as a non-trivial and crucial element of SSL method transfer. In the second step, we set hyperparameters to their optimal values and test out the quality of the representations learned through each of these methods on two classic surgical downstream tasks: phase recognition and tool presence detection. \textcolor{changetext}{Furthermore, we} verify how these approaches respond to varying amounts of labeled and unlabeled data in a practical semi-supervised setting. Here, we show that these methods, while generic in design, achieve state-of-the-art performance for both tasks and significantly mitigate the reliance on annotated data, adding up to {\color{changetext}7.4\%} phase recognition $F_{1}$ score and 20.4\% tool presence detection mAP. {\color{newtext} In the final step of the study, we extend our experiments to additional tasks and datasets: phase recognition \& tool presence detection on HeiChole \citep{heichole}, phase recognition \& tool presence detection on CATARACTS \citep{al2019cataracts}, action triplet recognition with CholecT50 \citep{rdv}, semantic segmentation on Endoscapes \citep{endoscapes}, and 8 \& 25 class semantic segmentation with CaDIS \citep{cadis}; thereby extensively covering the domain of surgical vision with SSL.}

This paper's contributions are as follows:

\begin{enumerate}
    \item Benchmarking of four state-of-the-art self-supervised learning methods (MoCo v2 \citep{chen2020improved}, SimCLR \citep{chen2020simple}, SwAV \citep{caron2020unsupervised}, and DINO \citep{caron2021emerging}) in the surgical domain.

    \item Thorough experimentation (
    $\sim$200 experiments, 7000 GPU hours) and analysis of different design settings - data augmentations, batch size, training duration, frame rate, and initialization - highlighting a need for and intuitions towards designing principled approaches for domain transfer of SSL methods.

    \item In-depth analysis on the adaptation of these methods, originally developed using other datasets and tasks, to the surgical domain with a comprehensive set of evaluation protocols, {\color{changetext} spanning 10 surgical vision tasks in total performed on 6 datasets}.
    
    \item Extensive evaluation ($\sim$280 experiments, 2000 GPU hours) of the scalability of these methods to various amounts of labeled and unlabeled data through an exploration of both fully and semi-supervised settings.
\end{enumerate}

\section{Related Work}
{\color{changetext} \subsection{Self-supervised representation learning in computer vision}}
\label{sec:ssl_general}
In the absence of external labels, SSL methods rely on the input image's intrinsic information to define a proxy loss to minimize. This artificial loss forces the model to learn rich vector representations of images, i.e. vectors in an embedding space with relative positions that meaningfully reflect the original visual content. The underlying expectation is that these representations are suitable for a wide range of useful downstream tasks.

The following paragraphs provide an overview of the various categories of SSL methods, tracing their evolution over the past few years. {\color{changetext} Here we focus on non-surgical visual tasks, considering mostly general computer vision works as well as a few others in medical image analysis.}

{\color{changetext} \noindent\textbf{Early heuristics-based methods.}}
Early SSL approaches aimed to learn representations by training models to solve a simple handcrafted task with some degree of relevance to the target task \citep{kim2018learning}.
These included predicting spatial context \citep{doersch2015unsupervised}, image rotation \citep{gidaris2018unsupervised}, artificial classes based on geometric transformations \citep{dosovitskiy2014discriminative}, and image patch arrangement \citep{noroozi2016unsupervised}.
Similarly, other works proposed reconstructing image regions \citep{pathak2016context} or colorization \citep{zhang2016colorful, zhang2017split}.
An exhaustive review of SSL methods based on pretext tasks is conducted in~\cite{jing2020self}.

\noindent\textbf{Contrastive methods.} 
More recently, contrastive learning methods have emerged as an alternative to handcrafted heuristics. These methods place less emphasis on the nature of the pretext task, instead focusing on controlling the relative position of features in the embedding space. They rely on generating positive and negative pairs of samples, which are then passed to a discriminative loss function to generate a training signal.

Early works attempted to generate such samples from within a single image using image patches \citep{10.5555/2968826.2968912, oord2018representation}; however, these methods failed to take advantage of relationships between different images.
Consequently, \cite{wu2018unsupervised} proposed the concept of a memory bank to store representations of many instances, which they leverage to impose an inter-instance discrimination objective.
\cite{he2020momentum} refined this idea with MoCo, using a momentum encoder rather than a memory bank to store representations, thereby enabling the sampling of many more instance pairs for the discrimination objective. An improved version with an additional projection head and more augmentations, MoCo v2, was later proposed by \cite{chen2020improved}.
Recently, \cite{chen2020simple} introduced SimCLR, a simpler framework outperforming many previous works \citep{oord2018representation, bachman2019learning, henaff2020data, tian2020contrastive, misra2020self} by using aggressive data augmentations to generate `positive pairs' for the discrimination objective.

{\color{newtext} Among SSL approaches, contrastive learning in particular has seen extensive use in research on medical image analysis in recent years. This form of pretraining has been employed to support many medical vision tasks: most commonly classification for diagnostic purposes \citep{chen2021uscl, ke2021histopathology, yang2021focal, xing2021categorical, dong2021fed, zhao2021radiomics, huang2021retinopathy, dufumier2021proxy}, but also more complex tasks such as detection \citep{li2021domain, tian2021anomaly, lei2021localization}, segmentation \citep{wu2021fedseg, hu2021semi, zeng2021volumetric, boutillon2021pediatric, zhou2021acseg} and multimodal tasks combining text with vision \citep{liu2021vqa, jiao2020vsus}. Several imaging modalities are represented as well: MRI \citep{wu2021fedseg, hu2021semi, dufumier2021proxy, boutillon2021pediatric}, CT \citep{yang2021focal, lei2021localization, zhou2021acseg}, X-Ray \citep{li2021domain, liu2021vqa} and ultrasound \citep{chen2021uscl, jiao2020vsus}. }
\\
\noindent\textbf{Cluster-based and distillation-based methods.} 
While contrastive methods have brought significant performance improvements, requiring positive and negative sampling during training can be impractical, and has pushed the community towards alternative approaches.

Self-supervised clustering methods \citep{caron2018deep, asano2019self, caron2020unsupervised, grill2020bootstrap, caron2021emerging} provide another alternative to the pretext task-based approach, focusing on clustering latent image representations in embedding space.
Initially, Caron et. al. introduced DEEPCLUSTER \citep{caron2018deep}, which adapted the k-means algorithm to assign clusters to images.
\cite{asano2019self} showed reformulating cluster assignment as an optimal transport problem improves performance.
SwAV \citep{caron2020unsupervised} further improves on this by constraining augmented views of an image to have consistent cluster assignments.

Other works, based on distillation, bootstrap multiple neural networks in a teacher-student fashion to learn latent representations \citep{grill2020bootstrap}. DINO \citep{caron2021emerging} applies this bootstrapping approach with vision transformers, attaining state-of-the-art results.
\\
{\color{newtext} \noindent\textbf{Masked image modeling}. Techniques based on concealing parts of images, as mentioned in our previous paragraph on heuristics-based methods, have existed in the computer vision community for several years: \cite{pathak2016context}'s image region reconstruction is one early example of masked image modeling (MIM). The emergence of Transformer models, however, led to a resurgence of MIM. Drawing inspiration from masked language modeling tasks for Transformers in natural language processing, recently published masked image modeling techniques view images as sequences of visual tokens, representing patches in a grid. A selection of tokens in the sequence is masked, then prompted for prediction by a Transformer employing attention on the sequence's tokens.

iGPT \citep{chen2020iGPT} used a Transformer to predict individual pixels in images scaled down to low resolutions, while ViT \citep{dosovitskiy2020ViT} predicted the mean colors of masked patches. BEiT \citep{bao2022beit}, mc-BEiT \citep{li2022mcbeit}, and PeCo \citep{dong2021PeCo} learned to predict tokens produced by a VQ-VAE (Vector-Quantized Variational Auto-Encoder \citep{vq-vae}) from masked patches. MaskFeat \citep{wei2022MaskFeat} studied a broad spectrum of feature types and proposed to regress Histograms of Oriented Gradients (HOG) for the masked content. MAE \citep{he2022MAE} and SimMIM \citep{xie2022simmim} proceeded with direct regression on raw RGB pixel values.}
\\
\noindent\textbf{Spatio-temporal methods.}
Parallel to static image methods presented in the previous paragraphs, research on SSL has explored video data through approaches tailored to spatio-temporal models. Most of them rely on spatio-temporal heuristics, with more emphasis on timing \citep{vssl_shufflelearn, vssl_oddoneout, vssl_sorting, vssl_order, vssl_tcc, vssl_temporaltransform, vssl_speednet} or appearance \citep{vssl_colorization, vssl_jigsaw, vssl_move, vssl_cubic, vssl_dynamonet}. A few contrastive methods exist as well \citep{vssl_contrastive, vssl_videomoco, vssl_cotraining}. Recently, a large-scale study by \cite{vssl_review} adapted four single-frame SSL methods \cite{chen2020simple, he2020momentum, Byol2020, caron2020unsupervised} to video data and compared their performance.
\\

\noindent\textit{Position of our work. }
Self-Supervised Learning is an intensely active research topic, with a large number of very distinct approaches proposed in recent years. For this reason, choosing an SSL method - especially for anything other than natural image data - is a complex problem: comparisons presented in SSL works can only cover a small selection of methods. More importantly, these comparisons are mainly conducted on natural image datasets such as the Imagenet dataset \cite{imagenet}; no reference point exists for surgical datasets, which are entirely different in terms of appearance. This is precisely the gap we fill with our work: we study how SSL adapts to surgical computer vision using a choice of methods that sufficiently span the state-of-the-art for static images {\color{newtext} with methods based on contrastive learning, clustering, and distillation. Masked Image Modeling methods have not been selected since the patch division process that makes those suitable for Transformers would first need to be ported to the more classical architecture of ResNet50 (retained due to its status as the standard for SSL). This port alone would require extensive and dedicated experimentation.} Spatio-temporal models, while potentially relevant for future studies, are {\color{newtext} also} omitted here due to challenging and radically different temporal modeling requirements in the surgical domain: commonly used natural video datasets in SSL \citep{kinetics, ucf, hmdb} contain short clips of a single action, contrasting heavily with full recordings of surgical interventions.

\subsection{Surgical computer vision. }
General computer vision focuses on natural images with scenes and items from everyday life. In contrast, surgical computer vision aims at identifying surgical activities and objects with varying degrees of detail. Early work in the field focused on automatically recognizing surgical workflow at the coarsest level through two fundamental tasks: phase recognition and tool presence detection. These highly specialized visual tasks prompted developments in terms of methodology separately from the rest of computer vision, which we cover in the next paragraphs.

\noindent\textbf{Full supervision.}
Initial efforts in surgical computer vision involved phase recognition based on handcrafted features \citep{padoy_phase, blum_phase}. Deep learning was first introduced to the field by \cite{twinanda2016endonet} and \cite{dergachyova_phase}, replacing handcrafted features with embeddings extracted by convolutional neural networks; \cite{twinanda2016endonet} in particular introduced the \textit{Cholec80} dataset, containing 80 videos of cholecystectomy annotated with surgical phases and tool presence labels. This dataset has since remained as one of the {\color{newtext}surgical} computer vision community's main datasets \citep{sds}, appearing in most works mentioned in this paragraph.
With surgical workflow and continuity of surgical actions playing a major role in these tasks, spatio-temporal models quickly emerged, outperforming single-frame models by a wide margin. \cite{endolstm} employed combinations of CNNs and LSTMs for surgical phase recognition and tool presence detection. Since then, increasingly refined spatio-temporal architectures have been proposed to better model the tasks \textcolor{changetext}{\citep{svrcnet, mtrcnet, Czempiel2020TeCNOSP, TMRNet, opera}}. \textcolor{newtext}{Recently, \cite{BNpitfalls} studied end-to-end spatio-temporal models and the effect of Batch Normalization on the success of these models}. Outside of these examples, a more comprehensive overview of surgical phase recognition approaches is provided in a survey by \cite{phase_review}. For recognizing tools in cataract surgery, \cite{al2018monitoring} proposed combinations of CNNs and RNNs with boosting.
\\
\noindent\textbf{Self-supervision in surgery. }
Self-supervision is still in the very early stages of research within surgical computer vision. While SSL methods in general computer vision have evolved towards methods such as contrastive learning, clustering or distillation (Section \ref{sec:ssl_general}), self-supervision on surgical data is still mostly limited to heuristics; for instance, \cite{ross2018exploiting} uses a colorization pretext task. Furthermore, the self-supervised tasks seen in surgery generally involve external information: \cite{da2019self, sestini2021kinematic} incorporate robot kinematics. \cite{yengera2018less} rely on remaining surgery duration estimation as the pretext task to improve surgical phase recognition on Cholec80. The only existing examples of contrastive learning add external information as well: \cite{bodenstedt_framesort} used a frame sorting task; later, \cite{funke2018temporal} introduced a method named second-order temporal coherence. In both cases, comparisons between frames are driven by time (i.e. relative positions of frames inside of a video) instead of their actual content.
\\
\noindent\textit{Position of our work. } Current research on surgical computer vision heavily leans towards fully supervised methods, which require large amounts of data to be annotated with clinical expertise. For improved scalability, a few approaches involving self-supervision have been developed. These approaches, however, heavily rely on heuristics and external information; as such, they lag behind general SSL, which has expanded to a larger spectrum of methods in recent years, all purely based on pixel data. Our work targets this deficit by bringing recently proposed SSL methods to surgery and adapting them to this particular domain. Since single-frame feature extractors play a fundamental role in state-of-the-art spatio-temporal models in surgical computer vision, examining SSL methods designed for static images is an obligatory first step, which is the focus of this study.

\section{Methodology}

We first establish the setting of this study by introducing the relevant surgical data and tasks, followed by our selection of SSL methods. We then outline our experiments; {\color{changetext} three main stages are defined as shown in Fig. \ref{fig:concept}, the \emph{hyperparameter study} (A), the \emph{data supply study} (B) and the \emph{generalization experiments} (C). Stages A and B each examine in detail the reaction of SSL in the surgical domain to a different factor, respectively parameterization and available data quantities. Stage C is an extension of our experiments to a much larger variety of datasets and tasks.} Implementation details for each stage of this study are available in the supplementary material.

\subsection{Surgical data \& surgical tasks}

\noindent\textit{\textbf{Cholec80. }} Since its introduction by \cite{twinanda2016endonet}, the Cholec80 dataset has been the foundation for many studies in surgical computer vision; we, therefore, use it here for our SSL benchmark. This dataset contains 80 videos of complete laparoscopic cholecystectomy procedures, recorded at 25 frames per second with a resolution of 854 $\times$ 480 \textcolor{newtext}{or $1920 \times 1080$}. The average video duration is 38 minutes with 16 minutes of standard deviation, indicating a high degree of heterogeneity.

The two tasks used as downstream tasks are \textit{tool presence detection} and \textit{surgical phase recognition}, mirroring the \textit{object detection} and \textit{action recognition} tasks of general computer vision, respectively.

\begin{figure}[ht]
\centering
\includegraphics[width=\linewidth]{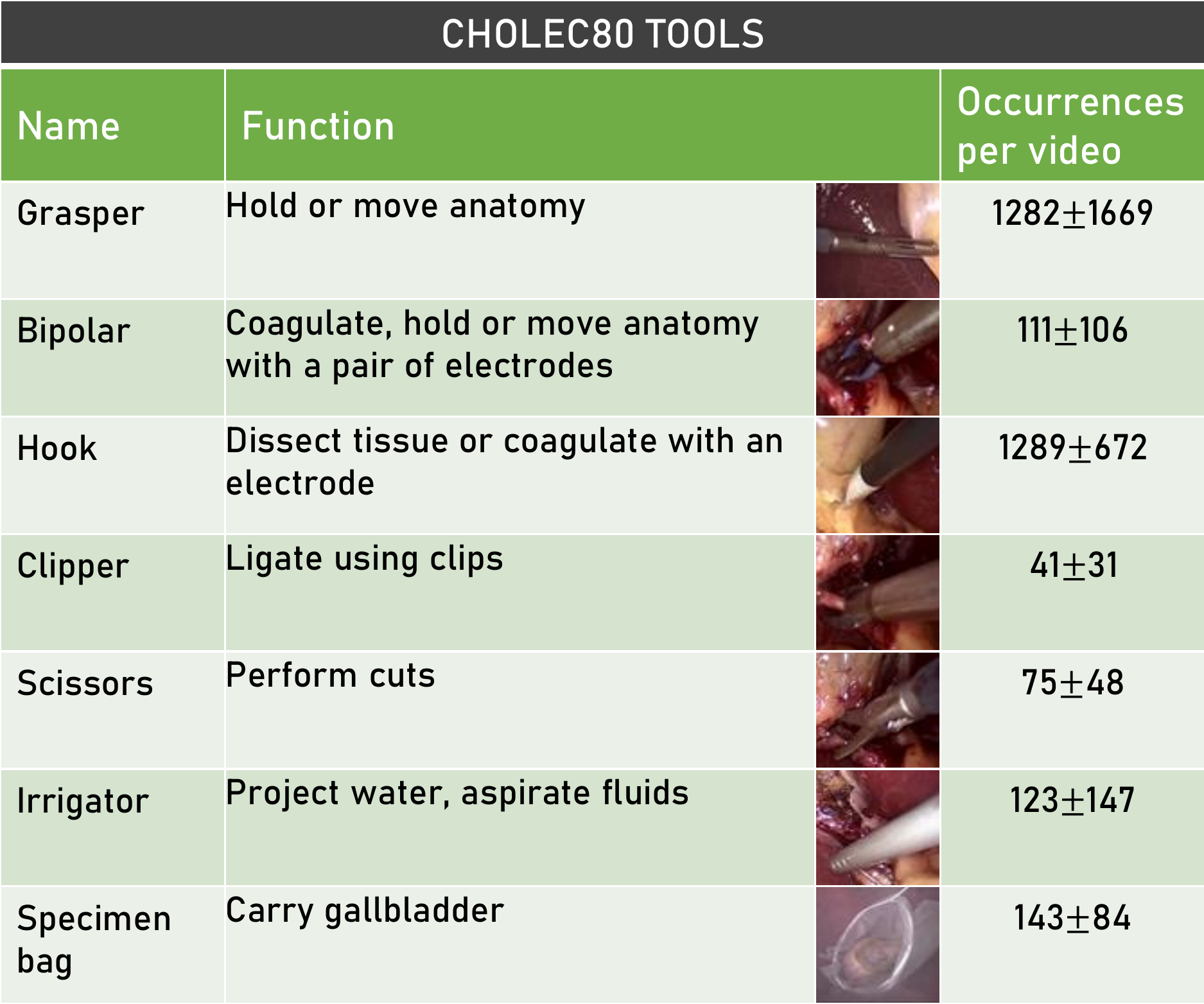}
\caption{Tools featured in the Cholec80 dataset.}
\label{fig:tool_types}
\end{figure}

\textit{Tool presence detection} is a multi-class, multi-label classification problem aimed at identifying all the surgical tools appearing in a given frame \citep{twinanda2016endonet,nwoye2019weakly,al2018monitoring}. It goes beyond image-level classification as zero, one, or several types of tools can be detected in one surgical image frame at the same time. 7 tools are featured, as described in Fig. \ref{fig:tool_types}.

\begin{figure}[ht]
\includegraphics[width=1.0\linewidth]{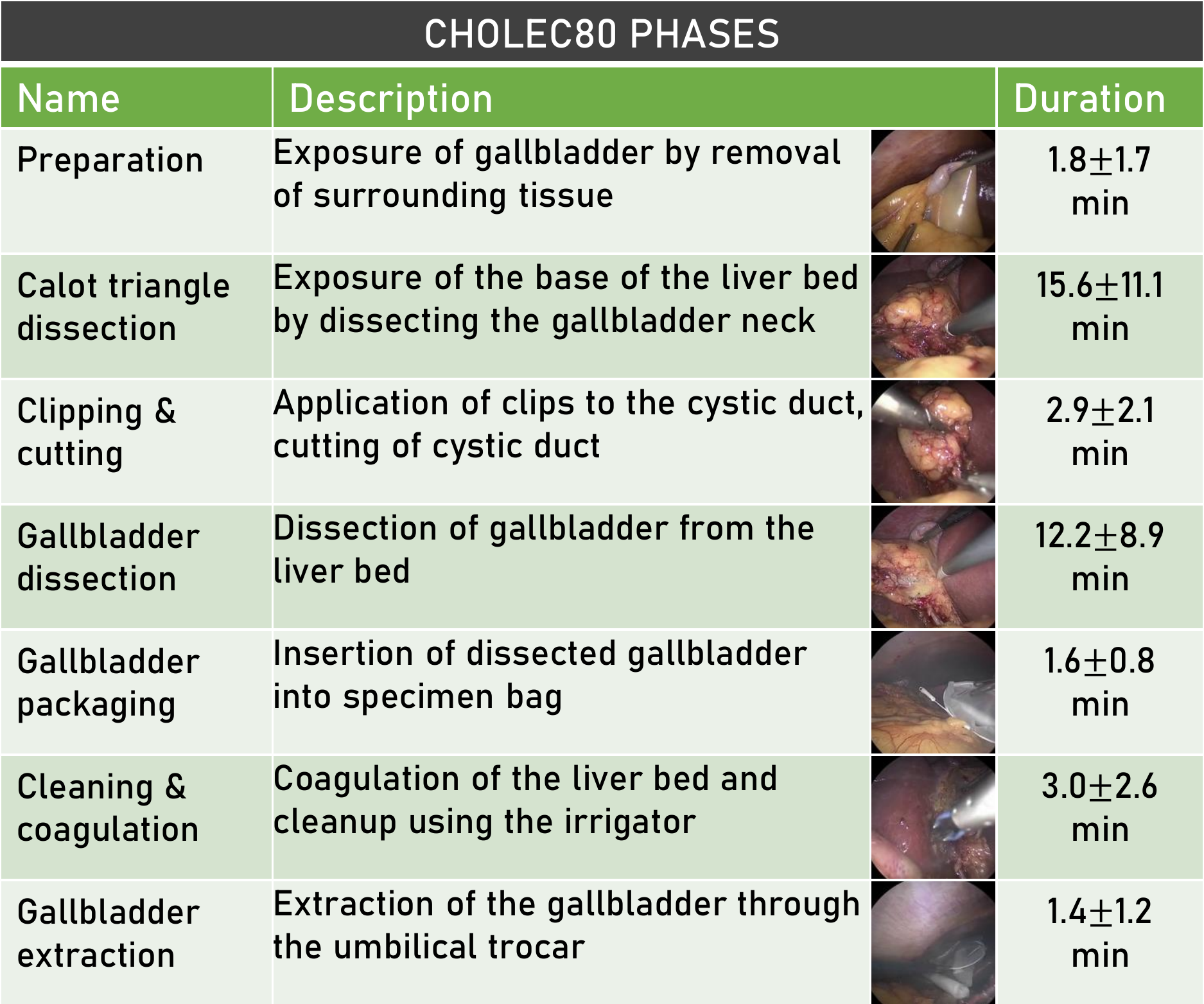}
\caption{Phases featured in the Cholec80 dataset.}
\label{fig:phase_table}
\end{figure}

\textit{Surgical phase recognition} entails classifying every frame of a recorded surgical procedure based on the activity being performed. This is a challenging task since important tools or anatomical parts often exit the field of view; as a result, useful visual indicators for making predictions tend to be quite sparse. Each procedure is decomposed into up to 7 phases described in Fig. \ref{fig:phase_table}.
\\
\textcolor{newtext}{\noindent{\textbf{Additional data \& tasks.}} While experiments featured in this work mostly focus on Cholec80 due to its prevalence in the community, a later stage of our study looks at other interesting datasets and surgical tasks. The digest of all datasets and tasks are presented in Fig. \ref{fig:additional_datasets}. 
\\
\noindent{\textbf{HeiChole.}} The HeiChole\footnote{\url{https://www.synapse.org/\#!Synapse:syn18824884/wiki/591922}} \citep{heichole} dataset, introduced as part of the EndoVis 2019 challenge, consists of 33 video recordings of cholecystectomy surgeries from three different hospitals. The training set, consisting of 24 videos, is publicly available while a test set of 9 videos is privately held for evaluation. The complete dataset contains frame-wise annotations of surgical phase and tool presence. Each procedure is segmented into 7 phases and could feature up to 7 tools. The description of all the phases and tools is presented in \cite{heichole}.
\\
\noindent{\textbf{CATARACTS.}} The CATARACTS dataset, introduced as part of the Challenge on Automatic Tool Annotation for cataRACT Surgery (CATARACTS)\footnote{https://cataracts.grand-challenge.org/} in 2017, is another popular dataset in the surgical vision community. The dataset consists of 50 recordings of cataract surgical procedures. In a recent edition of the challenge\footnote{https://www.synapse.org/\#!Synapse:syn21680292/wiki/601561} \citep{al2019cataracts}, the dataset was fully annotated for both tool presence detection and surgical activity recognition (step) tasks. In total, there are 19 steps and 21 different tool classes. We use the same splits as the CATARACTS 2020 challenge where the dataset was separated into 25, 5, and 20 videos corresponding to a train, validation, and test set, respectively.
\\
\noindent{\textbf{CholecT50.}} CholecT50 is a video dataset of laparoscopic cholecystectomy surgery introduced by \cite{rdv} to enable research on fine-grained action recognition. A collection of 50 videos, of which 45 videos are from the Cholec80 dataset and an additional 5 videos from an in-house dataset for cholecystectomy surgery, are fully annotated with action triplet information in the form of \textit{\textlangle{instrument, verb, target}\textrangle}. A total of 100 actions triplet classes are defined by \cite{rdv} as various combinations of 6 instruments, 10 verbs, and 15 targets. The dataset is split into 45 videos for training and 5 videos for testing, following the split used in the CholecTriplet2021 Challenge \footnote{https://cholectriplet2021.grand-challenge.org/}.
\\
\noindent{\textbf{Endoscapes.}} Introduced by \cite{endoscapes}, Endoscapes is a dataset comprised of 2208 frames selected at regular intervals (every 30 seconds) from 201 laparoscopic cholecystectomy videos with pixel-wise annotations for the task of semantic segmentation. A total of 29 semantic classes are defined in \cite{endoscapes} with 6 anatomy classes, 19 instrument classes, and 4 other miscellaneous classes. We follow the same data splits of \cite{endoscapes} in all our experiments.
\noindent{\textbf{CaDIS.}} CaDIS \citep{cadis} is a semantic segmentation dataset for cataract surgery. The dataset consists of 4670 images extracted extending part of the CATARACTS dataset with pixel-level annotations for 36 classes (29 surgical instrument classes, 4 anatomy classes, and 3 miscellaneous classes). The 4670 images are split into train, validation, and test sets comprising 3550, 534, and 586 images, respectively. Out of the three different evaluation tasks, representing increasing degrees of granularity, we consider the two extremes for evaluation in this study. Task I aims at differentiating anatomy and instruments in each frame and hence consists of 8 semantic classes: 4 classes for anatomical structures, 1 class for all instruments, and 3 classes for all other objects appearing in the images. Task III, on the other hand, focuses on more detailed instrument classification by representing each instrument type and instrument tips as separate classes totaling 25 classes.}
\\
\begin{figure*}[t!]
\includegraphics[width=1.0\linewidth]{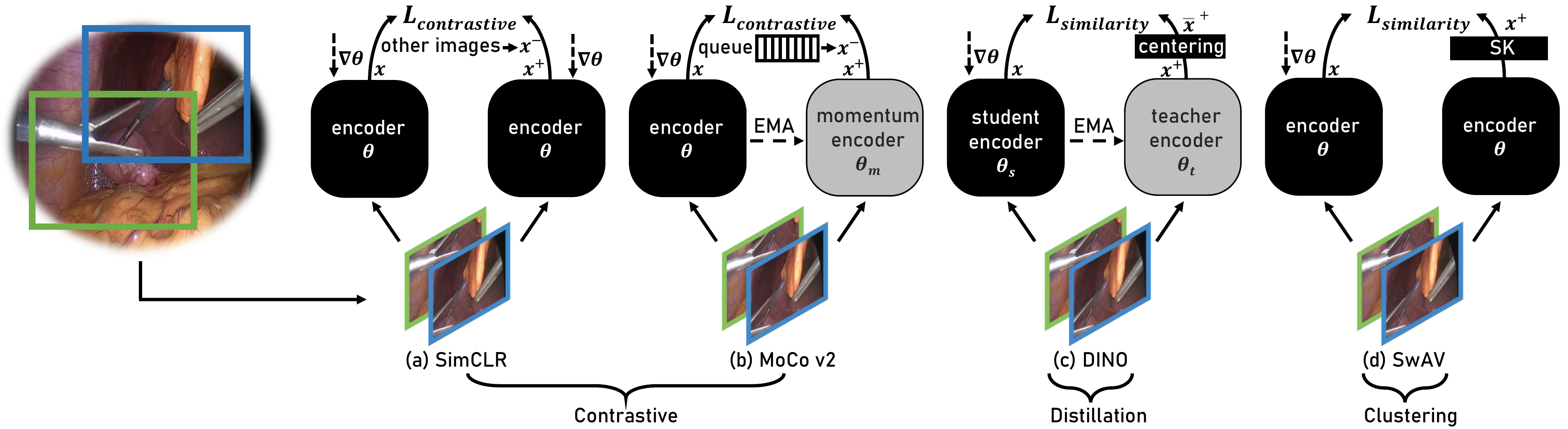}
\caption{We study four SSL methods from three categories: contrastive (SimCLR \citep{chen2020simple} and MoCo v2 \citep{he2020momentum,chen2020improved}), distillation-based (DINO \citep{caron2021emerging}), and clustering-based (SwAV \citep{caron2020unsupervised}). SimCLR and MoCo v2, as contrastive methods, use embeddings from other images or a queue to generate negative embeddings ($x^-$), respectively. MoCo v2 and DINO use an explicit momentum encoder whose weights are updated using an exponential moving average (EMA). $\nabla \theta$ are the gradients of the encoder's weights $\theta$, computed using a contrastive loss ($L_{contrastive}$) for SimCLR and MoCo v2 and a similarity loss ($L_{similarity}$) for DINO and SwAV. DINO uses a centering operation, and SwAV uses a non-differentiable Sinkhorn-Knop (SK) transform \citep{Cuturi13} to avoid mode collapse in the absence of negative embeddings.}
\label{fig:intro-methods}
\end{figure*}
\subsection{Selected SSL methods}
As shown in Section \ref{sec:ssl_general}, general computer vision offers a wide range of SSL methods. In order to adequately represent the current state of the art, we select a total of four SSL methods: two contrastive (SimCLR \citep{chen2020simple}, MoCo v2 \citep{he2020momentum,chen2020improved}), one distillation-based (DINO \citep{caron2020unsupervised}), and one clustering-based (SwAV \citep{caron2020unsupervised}), see Fig. \ref{fig:intro-methods}.

Several studies on unsupervised visual representation have proposed approaches based on contrastive learning \citep{hadsell2006dimensionality,wu2018unsupervised,van2018representation,hjelm2018learning,zhuang2019local,henaff2020data,tian2020contrastive,bachman2019learning}, with the core idea being to maximize the representational similarity for pairs of positive samples and dissimilarity for pairs of negative samples. A key component of these methods is mining positive and negative samples in a batch without explicit labels. A common approach in these methods is, for each image, to consider its augmentations as a corresponding positive sample, and other images as corresponding negative samples. The positive and the negative samples are passed through a base encoder to obtain the corresponding positive ($x$, $x^+$) and negative ($x^-$) embeddings. The InfoNCE loss \citep{oord2018representation} commonly used in contrastive methods is defined as follows:
\begin{equation}
   L_{contrastive} = \mathbb{E}_{x,x^+,x^-}\left[-\log\frac{e^{x \cdot x^+ / \tau}} {e^{x \cdot x^+ / \tau} + (\sum_{k=1}^{K}e^{x \cdot x^- / \tau})}\right], 
\end{equation}
where $\tau$ is a temperature hyperparameter for scaling the embeddings. The negative samples are required in contrastive methods to avoid model collapse to an identity solution. Each of the following four selected SSL methods works on similar principles with a few modifications.

\noindent\textbf{SimCLR} \citep{chen2020simple} considers the other images from a batch as negative samples and passes them through the encoder to obtain the negative embeddings ($x^-$) to compute the contrastive loss, $L_{contrastive}$, using equation (1).

\noindent\textbf{MoCo v2} \cite{he2020momentum} introduced MoCo, employing a large memory queue to store negative embeddings $x^-$. This queue allows decoupling the dictionary size from the mini-batch size, in order to perform well even with smaller batch sizes. Furthermore, since the queue contains embeddings from different mini-batches, a momentum encoder is used to enforce consistency across different mini-batches. The weights of the momentum encoder ($\theta_m$) are updated using an exponential moving average (EMA) of the weights of the encoder ($\theta$): $\theta_m = \lambda\theta_m  + (1 - \lambda) \theta$, where $\lambda$ is a decay parameter. MoCo v2 \citep{chen2020improved} refines this design using an additional projection head and more augmentations.

\noindent\textbf{DINO} \citep{caron2021emerging}, inspired by BYOL \citep{Byol2020}, uses a teacher-student approach in a knowledge-distillation framework \citep{hinton2015distilling}. The student encoder, parameterized by $\theta_s$, and the teacher encoder, parameterized by $\theta_t$, are used to generate two positive embeddings, $x$ and $x^+$, respectively. Similar to MoCo v2, the weights of the teacher encoder are updated using EMA. However, DINO also removes the dependency on negative samples; in the absence of negative embeddings, this method avoids \emph{model collapse} using a \emph{centering} operation. This operation first computes the centers of the positive embeddings using EMA, $c = \lambda_cc + (1-\lambda_c) \frac{1}{B}\sum\limits_{i=1}^{B}x_i^+$, then subtracts the centers $c$ from the positive embeddings to compute the mean-centered positive embeddings, $\bar{x}^+ = x^+ - c$. Here, $B$ is a batch dimension and $\lambda_c$ is a centering decay parameter. The similarity loss 
\begin{equation}
L_{similarity} = -\sum \text{softmax}(x/\tau_s)\log(\text{softmax}(\bar{x}^+/\tau_t)) 
\end{equation}
is computed as a cross-entropy loss between the reference positive embedding, $x$, and mean-centered positive embeddings, $\bar{x}^+$. The $\text{softmax}()$ function normalizes embeddings that are scaled differently using temperature parameters $\tau_s$ and $\tau_t$ for the student and teacher encoders, respectively.

\noindent\textbf{SwAV} \citep{caron2020unsupervised} circumvents the need for negative embeddings by first transforming the positive embedding pair, $x$ and $x^+$, to learned prototype embeddings, $\bar{x}$ and $\bar{x}^{+}$ and then performing online clustering of the learned prototype embeddings using the Sinkhorn-Knopp (SK) algorithm \citep{Cuturi13}. The SwAV similarity loss is 
\begin{equation}
 L_{similarity} = \mathcal{D}_{KL}(\bar{x} \parallel \text{SK}(\bar{x}^{+})),
\end{equation}
where $\mathcal{D}_{KL}$ is the Kullback-Leibler divergence.

\subsection{Hyperparameter study design}\label{sec:hparam_study}
In the hyperparameter study (Fig. \ref{fig:concept}, A), we aim to better understand the sensitivity of each SSL method to hyperparameter variations and establish a set of \textbf{recommended} values that will later serve in practical use cases of semi-supervised learning, as part of the data supply study (Fig. \ref{fig:concept}, B). To this end, we select a subset of 5 critical hyperparameters:

\begin{itemize}
    \item Type of augmentation
    \item Batch size
    \item Epochs
    \item Sampling rate
    \item Type of initialization
\end{itemize}

We then carefully analyze the influence of all 5 on the model performance, for the tasks of phase recognition and tool presence detection on the Cholec80 dataset. Each of those 5 hyperparameters defines a group of experiments, where the relevant hyperparameter varies while others are set to the default values shown in Table \ref{tab:hp_defaults}. For each value of that hyperparameter, 4 models are trained - one for each selected SSL method. Linear evaluation is then performed on the validation set, i.e. by training a linear classifier added on top of the frozen backbone layers, for tool and phase tasks separately. This validation protocol, commonly used in SSL \citep{feichtenhofer2021large}, verifies here how well each method, for that particular hyperparameter value, maps frames to linearly separable vector representations that are consistent in terms of phase and tool content.
\begin{table}[]
\centering
\small
\caption{Observed SSL hyperparameters. Defaults are used in the hyperparameter study. Recommended values (best overall performance in the hyperparameter study) are used in the data supply study.}
\label{tab:hp_defaults}
\begin{tabular}{llcc}\hline
\multicolumn{2}{l}{}                                   & \textbf{Defaults} & \textbf{Recommended} \\ \hline
\multirow{4}{*}{\textbf{Augmentations}} & Multi-Crop   & 8        & 2                         \\
                                        & Color        & On       & On                        \\
                                        & Geometric    & On       & On                        \\
                                        & Strong-color & Off      & Off                       \\
\multicolumn{2}{l}{\textbf{Batch size}}                & 512      & 256                       \\
\multicolumn{2}{l}{\textbf{Epochs}}               & 300      & 300                       \\
\multicolumn{2}{l}{\textbf{Sampling rate}}       & 1        & 5                         \\
\multicolumn{2}{l}{\textbf{Initialization}}            & Scratch  & Imagenet                  \\
& & & fully supervised\\\hline
\end{tabular}
\end{table}
Details for each experiment group are provided in the following paragraphs.

\begin{figure}[ht]
\includegraphics[width=1.0\linewidth]{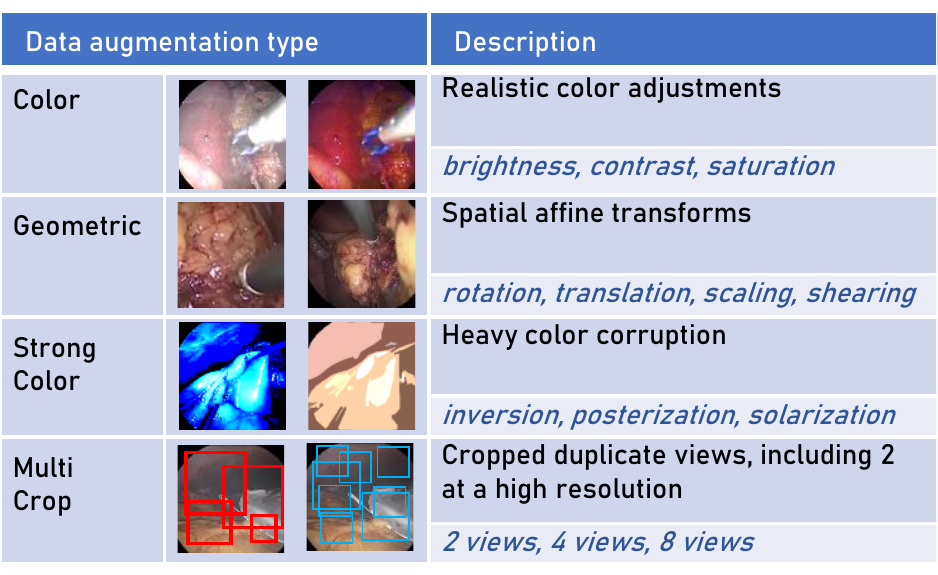}
\caption{Data augmentation types involved in the hyperparameter study}
\label{fig:aug_types}
\end{figure}

\noindent\textbf{Augmentations. }
Data augmentation is a crucial aspect of SSL methods \citep{chen2020simple}: learning persistent feature representations between different \textit{views} of the same image (i.e. between different augmented versions of the original image), is the implicit task that SSL methods leverage in order to produce powerful representations of unlabeled data.
Hence, it is imperative to understand the impact of this parameter when shifting to different domains and tasks. While an exhaustive search of augmentations is beyond the scope of this study \footnote{Pretraining a ResNet-50 using SSL with a single hyperparameter setting given our experimental design demands approximately 40 GPU hours using 4 NVIDIA V100s on average across considered methods.}, we decided to focus on broad categories of commonly used augmentation techniques to train SSL methods \citep{caron2021emerging,chen2020simple,he2020momentum}, defined here as \textit{Color}, \textit{Geometric}, \textit{Strong-color} and \textit{Multi-Crop}. Fig. \ref{fig:aug_types} provides a description for each category.

All the mentioned augmentations are randomized during training \citep{randaugment}; the randomization process follows the implementation of \cite{goyal2021self}.

\textit{Multi-Crop} is set to 2, 4, or 8 crops with 2 crops always sampled at a high resolution following \cite{caron2020unsupervised}. Each of the other 3 augmentation types is either \textit{on} or \textit{off}. Considering all the possible combinations, we examine a total of $3*2^{3}=24$ configurations for augmentations.
\\
\noindent\textbf{Batch size. }
Batch size is a crucial hyperparameter in SSL methods: SimCLR \citep{chen2020simple} established a positive correlation between performance and batch size attributed to the size of the pool of negative samples to draw from during training. The other 3 approaches have presented the ability to better function with smaller batches as an advantage, cutting down memory requirements. 

To examine these claims, we use batches of sizes 128, 256, 512, and 1024.
\\
\noindent\textbf{Epochs.}
Previous studies have shown that training time could largely impact SSL performance. Given this, we investigate the impact of training time by training each SSL method for 50, 100, 200, and 300 epochs.
\\
\noindent\textbf{Sampling rate.}
While the SSL methods we test are designed for still images, we can apply them to video inputs by simply extracting individual frames from each video.
A key consideration when doing so is the frame sampling rate, as this can affect the relative homogeneity among various input images. 
In this aspect, surgical videos pose a particularly interesting technical setting, as they tend to provide a stable context, and the only changes across frames, even for several minutes of video, are manipulations of organs and medical tools in the field of view.  
Consequently, while increasing the number of frames sampled per second dramatically increases the available training data, it is unclear whether this additional data would be beneficial for SSL methods.

We experiment with sampling videos at 0.1, 0.33, 0.5, 1, 3, and 5 frames per second (fps). 

\subsection{Data supply study design}\label{sec:scalability_study}
In contrast with the previous section, the data supply study (Fig. \ref{fig:concept}, B) operates with a completely fixed set of recommended hyperparameters (Table \ref{tab:hp_defaults}), suitable for examining our chosen SSL methods in practical semi-supervision use cases: instead of freezing the backbone after self-supervised training, here we finetune it with phase or tool annotations in conjunction with a linear classifier. For phase recognition, we also observe the performance obtained by adding a temporal model (TCN, \cite{Czempiel2020TeCNOSP}) after this step and finetuning it separately as well: this provides a strong point of comparison against the state of the art, while also gauging the representations learned through SSL when used in a temporal context.
\\
\noindent\textbf{Labeled data supply. }
We first focus on labeled data only. Performance with respect to annotated data availability (Fig. \ref{fig:concept}, B1) is examined in three settings, with supervised finetuning performed after SSL on 40 videos (100\% of the entire Cholec80 training set), 10 videos (25\%), or 5 videos (12.5\%) of the full data. To mitigate the effect of outliers, experiments for the last two settings are replicated on 3 randomly selected sets of videos. In all these configurations, the same 40 unlabeled videos are used for self-supervised pretraining.
\\
\noindent\textbf{Unlabeled data supply. }
In addition to this core set of experiments focusing exclusively on varying labeled data, we select one SSL method - MoCo v2 - and examine how it reacts to changes in the amount of unlabeled data  (Fig. \ref{fig:concept}, B2) used for self-supervised training: from 1 to 10, 20, 40 and finally 80 unlabeled videos. Results are reported for varying numbers of labeled videos used for finetuning.

\begin{figure}[ht]
\centering
\includegraphics[width=\linewidth]{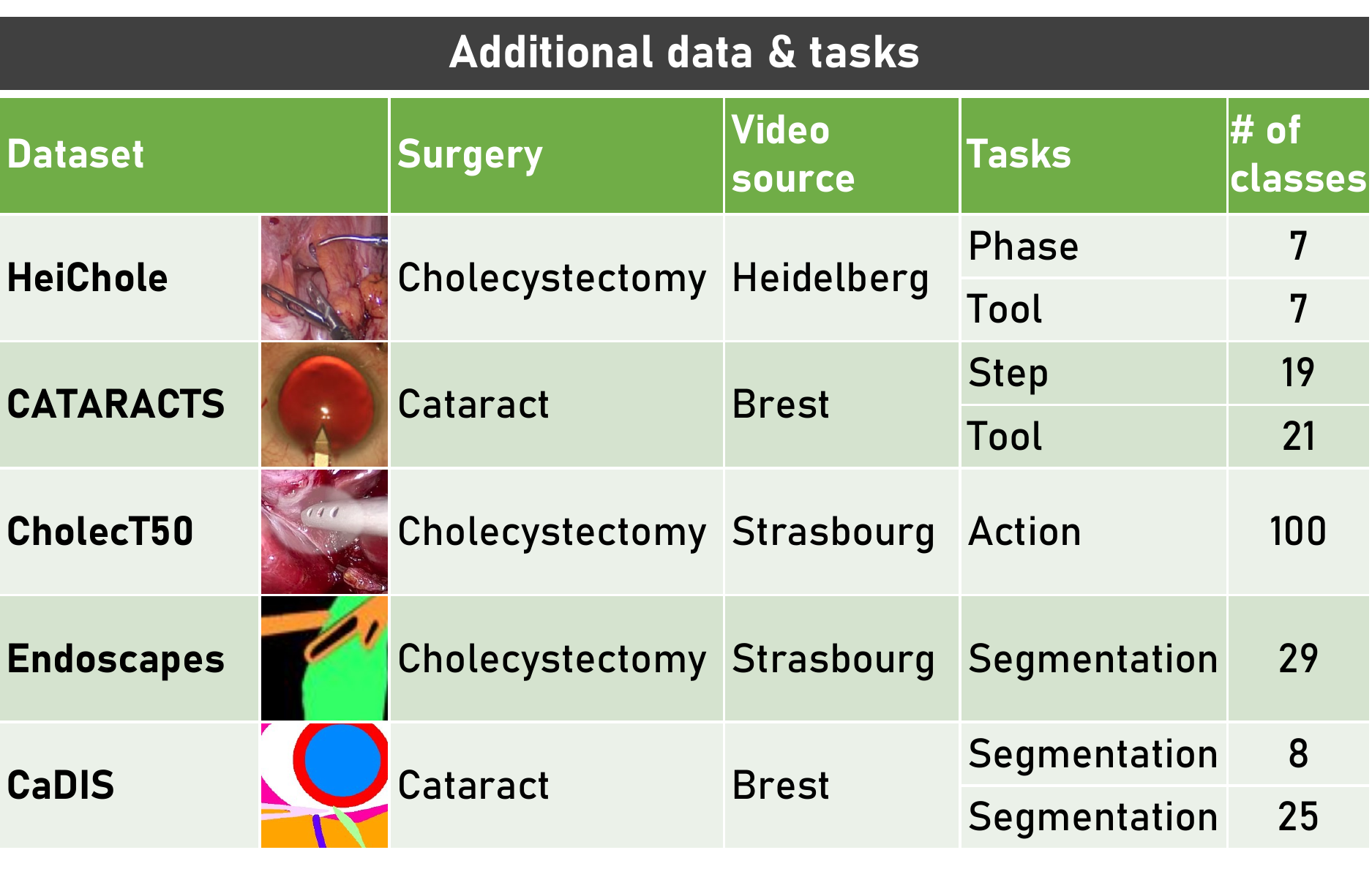}
\caption{{\color{newtext} Data featured in the generalization experiments.}}
\label{fig:additional_datasets}
\end{figure}

{\color{newtext} \subsection{Generalization study}
\label{sec:generalization}
\textcolor{changetext1}{Experiments conducted up to this point feature the Cholec80 dataset with two tasks - phase recognition and tool detection} \textcolor{newtext1}{- representing only a small portion of the variability of datasets used in surgical data science literature \citep{sds_2}}. In order to determine how well SSL generalizes to entirely different situations within surgery, we provide in this final stage a set of complementary experiments of a previously selected SSL method - MoCo v2 - inspecting its behavior across a total of 8 tasks across 5 different surgical datasets: HeiChole \citep{heichole}, CATARACTS \citep{Cataracts2018}, CholecT50 \citep{nwoye2019weakly}, Endoscapes \citep{endoscapes}, and CaDIS \citep{cadis}. Here the scope of the study is expanded by a considerable amount in several aspects. 
First, we study the effect of the SSL methods on the same surgical procedure and tasks but on diverse clinical centers, with surgical data sourced from 3 German hospitals (HeiChole).
Next, we investigate another type of minimally invasive surgery, i.e., cataract, through the CATARACTS dataset, offering a radically different visual appearance from cholecystectomy. Here again, we consider similar downstream tasks of surgical activity (step) recognition and tool presence detection. 
We further extend our analysis of SSL methods on yet another task, surgical action triplet recognition, on the recently released CholecT50 dataset. We add surgical scene segmentation as well with the Endoscapes dataset. 
Finally, we conclude the generalization study by analyzing the SSL methods on another surgical procedure and task with the CaDIS dataset for scene segmentation in cataract surgery.
A visual summary of the different dataset characteristics is shown in Fig. \ref{fig:additional_datasets}.}

\begin{figure*}[ht]
  \centering
  \begin{subfigure}
    \centering
    \includegraphics[width=0.47\linewidth]{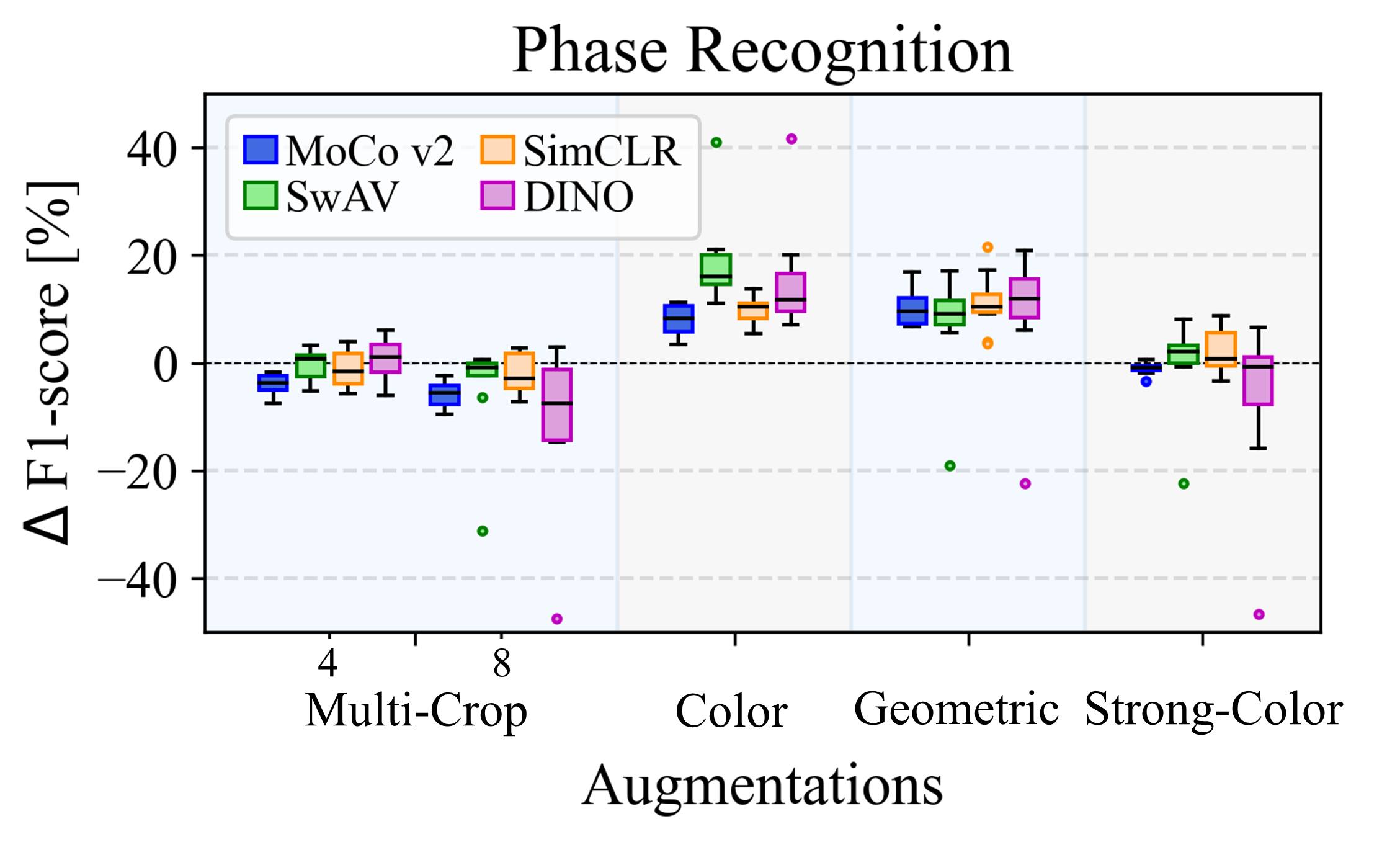}
  \end{subfigure}
  \begin{subfigure}
    \centering
    \includegraphics[width=0.47\linewidth]{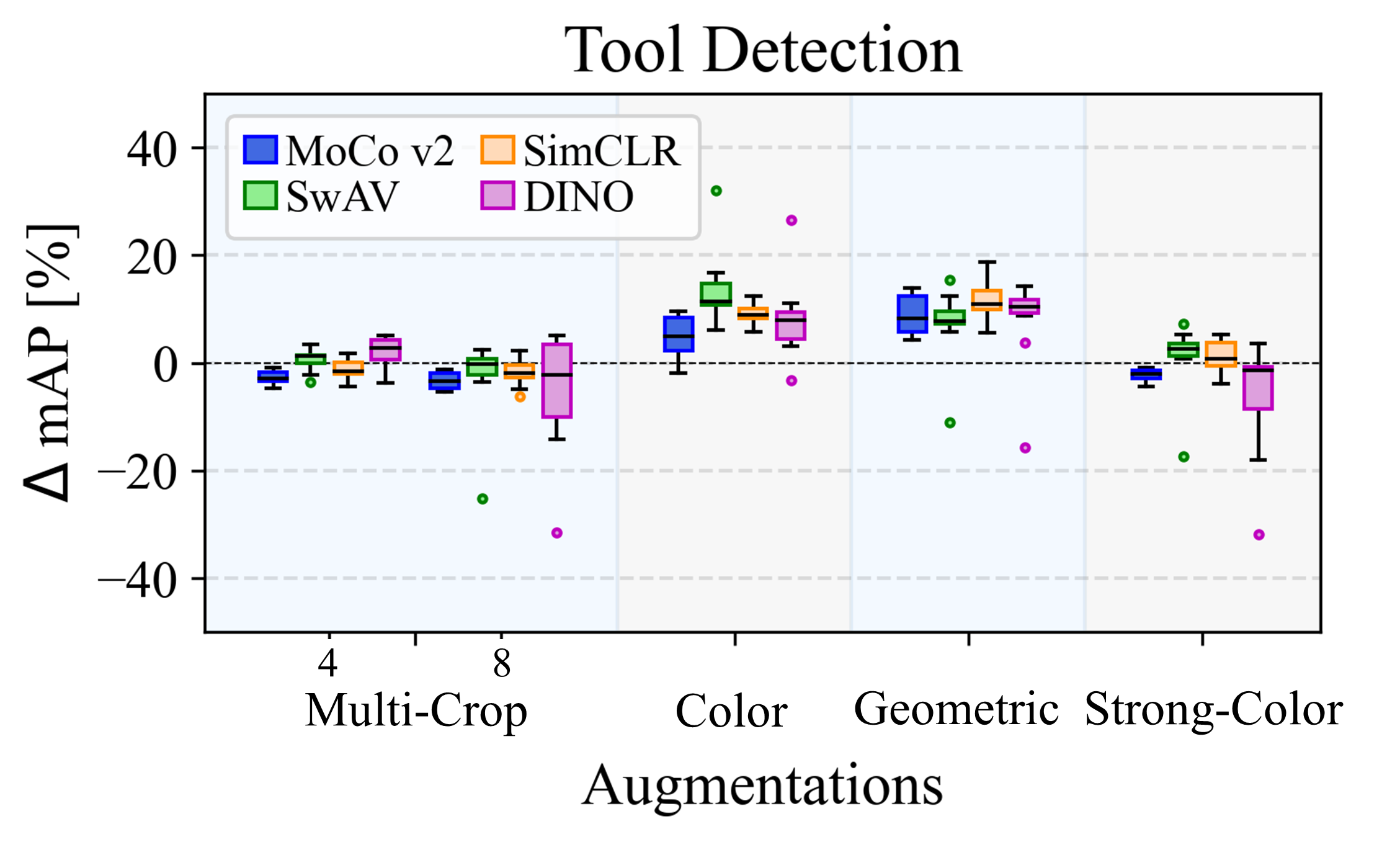} 
  \end{subfigure}
  \caption{Performance of each method on Cholec80 varying the augmentation strategy for self-supervised pretraining. For each method and category of augmentations, we show a boxplot with the change in performance from the default no-augmentation setting (using 2 crops for \textit{Multi-Crop}), by enabling that category of augmentation (using 4 or 8 crops for \textit{Multi-Crop}). The boxplot whiskers were set to 1.5 times the interquartile range beyond the first and third quartile; settings outside of this margin were defined as outliers and plotted as dots. Results were obtained using linear evaluation on the validation set. Left: $F_{1}$-score for phase recognition. Right: mAP for tool presence detection.}
  \label{fig:abl_augm}
\end{figure*}

\section{Results}
\subsection{Dataset Splits and Evaluation Metrics}
\label{sec:metrics_and_splits}
In all our experiments, following previous literature \citep{Czempiel2020TeCNOSP, svrcnet, twinanda2016endonet, opera}, we use 40, 8, and 32 videos from Cholec80 as our total available pool of training videos, our validation set, and our test set, respectively.

In the hyperparameter study, we perform SSL pretraining on the entire pool of 40 training videos and report the results on the validation set.

In the data supply study, we further conduct semi-supervised experiments with 5 videos (12.5\% of Cholec80 training set) and 10 videos (25\% of Cholec80 training set) of annotations, for which we employ two different sampling strategies. For the comparison with external methods (Table \ref{tab:cuhk}), we use the predefined dataset split introduced in \cite{shi_surgssl} as a sampling strategy to enable fair comparisons. However, for the remainder of our experiments (see Tables \ref{tab:fcn_main}, \ref{tab:tcn_main}, \ref{tab:tool_main}, and Figures \ref{fig:data_supply_phase}, \ref{fig:data_supply_tools}), we either make use of established training splits \citep{twinanda2016endonet} for larger data settings (40, 80 training videos), employ a stratified random sampling approach or random uniform sampling when stratifying is infeasible (1 training video). In each case when randomly sampling, we sample three separate subset splits of the training videos, evaluate model performance on each split, and report the mean and standard deviation across splits. Doing so alleviates selection bias and allows for sound comparisons across methods and experimental settings. Indeed, we find that the variance in performance across dataset splits, particularly in the low-data settings, can surpass performance differences across methods, highlighting the need to sample multiple splits.

For all phase {\color{changetext}and step} recognition experiments, with the exception of the external comparison (Table \ref{tab:cuhk}), we report per-video F1 Score, computed by averaging across each video's F1 score. In these tables, the standard deviation is presented across the sampled splits. Meanwhile, for the external comparison, we report a \textit{relaxed boundary} per-video $F_{1}$ Score, originally introduced in the m2cai16-workflow challenge \footnote{http://camma.u-strasbg.fr/m2cai2016/index.php/program-challenge/} and used by \cite{shi_surgssl}, to enable a fair comparison. The relaxed boundary metric introduces a 10 second `relaxed' period centered around each ground truth phase transition; during these periods, the two consecutive phases are considered to be correct classifications (e.g. phase 4 and phase 5 are both accurate classifications in the 10 seconds before and after the transition from phase 4 to 5). Consequently, the relaxed boundary metric results in higher scores across methods.

For all tool presence detection experiments, we compute mAP across all considered frames and in all the presented tables the standard deviation is calculated across splits. {\color{newtext} Action triplet recognition performance on the CholecT50 dataset is measured using mAP over the 100 valid triplet classes. Segmentation tasks featured in the generalization experiments are all evaluated using $F_{1}$ score.}

\subsection{Hyperparameter study}
\label{sec:hyperparameter_study}
We present here the impact of hyperparameters variations on the quality of the representations learned by the SSL methods we selected, following the setup described in Section \ref{sec:hparam_study} \footnote{GPU training presents some non-determinism that is not trivial to avoid. Because performing several reruns of every experiment in the hyperparameter study would be computationally impractical, we do so for one method selected at random and present the standard deviation when performing linear evaluation for both downstream tasks in order to contextualize our results. The standard deviation across 5 reruns for this selection for phase recognition and tool presence detection is 0.7 \% F1 and 0.7 \% mAP, respectively.}.

\begin{figure}[h!]
  \begin{subfigure}
    \centering
    \includegraphics[width=0.48\linewidth]{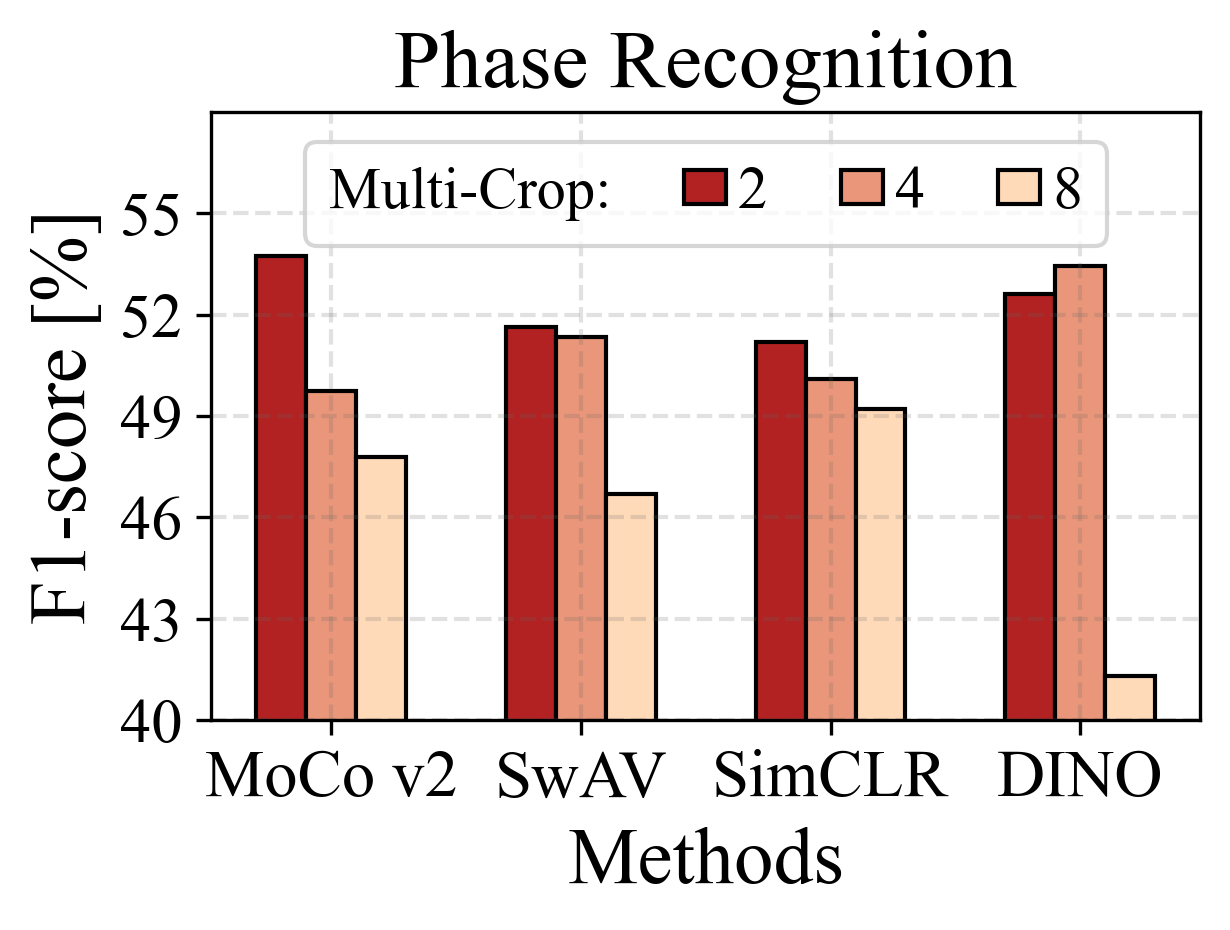}
  \end{subfigure}
  \begin{subfigure}
    \centering
    \includegraphics[width=0.48\linewidth]{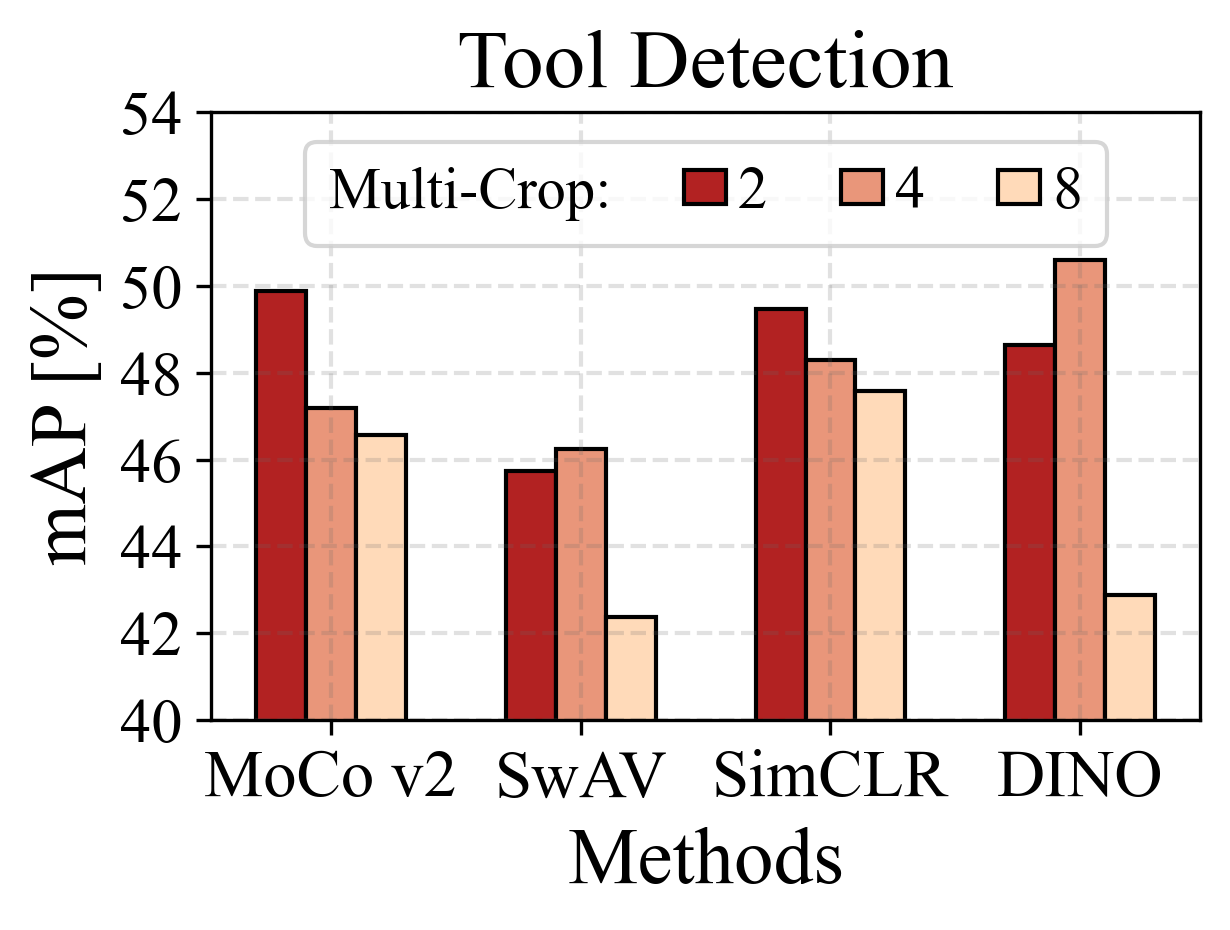}  
  \end{subfigure}
  \caption{Performance of each method on Cholec80 varying the Multi-Crop augmentation strategy for self-supervised pretraining: 2,4 or 8 crops (2 high-resolution crops, remaining low resolution). Results were obtained using linear evaluation on the validation set. Left: $F_{1}$-score for phase recognition. Right: mAP for tool presence detection.}
  \label{fig:abl_mc}t
\end{figure}

\noindent\textbf{Augmentations.}
In order to evaluate the impact of each of the four augmentation categories, we show the improvement introduced by the presence of each category across all the considered experiments for each SSL method. For every augmentation category, we examine the change in performance - $\Delta F_{1}$ and $\Delta mAP$ - caused by toggling it \textit{on} (for \textit{Multi-Crop}, by switching it from 2 to either 4 or 8). To this end, in Fig. \ref{fig:abl_augm}, we plot the following set of samples for the \textit{Multi-Crop} (4 and 8 crops - $\mathbf{MC4}$ and $\mathbf{MC8}$), \textit{Color} ($\mathbf{C}$), \textit{Geometric} ($\mathbf{G}$) and \textit{Strong-Color} ($\mathbf{S}$) augmentation experiments, respectively:
\begin{equation}
\begin{split}
    \mathbf{MC8} = \{ (mc_{\mathbf{8}\phantom{-}}c_{i}g_{j}s_{k}-mc_{\mathbf{2}}c_{i}g_{j}s_{k})_{i=\{1,0\}, j=\{1,0\}, k=\{1,0\}}\},\\
    \mathbf{MC4} = \{ (mc_{\mathbf{4}\phantom{-}}c_{i}g_{j}s_{k}-mc_{\mathbf{2}}c_{i}g_{j}s_{k})_{i=\{1,0\}, j=\{1,0\}, k=\{1,0\}}\},\\
    \mathbf{C} = \{ (mc_{i}c_{\mathbf{1}}g_{j}s_{k}-mc_{i}c_{\mathbf{0}}g_{j}s_{k})_{i=\{2,4,8\}, j=\{1,0\}, k=\{1,0\}}\},\\
    \mathbf{G} = \{ (mc_{i}c_{j}g_{\mathbf{1}}s_{k}-mc_{i}c_{j}g_{\mathbf{0}}s_{k})_{i=\{2,4,8\}, j=\{1,0\}, k=\{1,0\}}\},\\
    \mathbf{S} = \{ (mc_{i}c_{j}g_{k}s_{\mathbf{1}}-mc_{i}c_{j}g_{k}s_{\mathbf{0}})_{i=\{2,4,8\}, j=\{1,0\}, k=\{1,0\}}\},
\end{split}
\end{equation}

\noindent where $mc$ is \textit{Multi-Crop} augmentation and can take the values 2,4,8; $c$, $g$, $s$ are, respectively, \textit{Color}, \textit{Geometric} and \textit{Strong-Color} augmentations, which can either be toggled \textit{on} (1) or \textit{off} (0).
For each augmentation setting, statistics for $\Delta F_{1}$ and $\Delta mAP$ are collected and represented as boxplots. The average performance for each \textit{Multi-Crop} setting is also shown separately in Fig. \ref{fig:abl_mc}.

Experimental results for phase recognition and tool presence detection, shown in Fig. \ref{fig:abl_augm}, demonstrate the clear impact that augmentation strategies have on the quality of the learned representations, consistent across methods and tasks. We make three main observations:

(1) In general, increasing the number of low-resolution views on \textit{Multi-Crop} negatively impacts performance. From 2 crops for MoCo v2, switching to 4 crops cuts down phase recognition $F_{1}$ by 3.5\%; switching to 8 cuts it down by 4.5\%. This represents an important deviation from typical results in the natural image domain, where additional low-resolution views in \textit{Multi-Crop} generally positively correlated with improved performance \citep{caron2020unsupervised,caron2021emerging}. A possible explanation may be the weaker value of ensuring `local-to-global' feature invariance in the surgical domain; in surgical phase recognition, for example, discriminative cues may be scattered in the entire image, and be significant only if considered as a whole: in light of this, forcing `local-to-global' invariant features may be challenging, or even undesirable in this domain.

(2) The \textit{Color} augmentation consistently and significantly improves performance. This is generally analogous to results on the natural image domain \citep{feichtenhofer2021large}: as pointed out in \citep{chen2020simple}, augmentations like \textit{Multi-Crop} and \textit{Geometric} mostly preserve the original color distribution, leaving this as an easy shortcut for the network to solve the predictive task; the \textit{Color} augmentation is, therefore, an important factor in learning meaningful representations.

(3) DINO is the method most affected by the specific choice of augmentation; in particular, representation quality dramatically drops when both \textit{Multi-Crop} and \textit{Strong-Color} augmentations are used; a possible explanation may derive from the general observation on \textit{Multi-Crop} made previously: compared to the other methods, DINO explicitly enforces the `local-to-global' feature invariance by passing all views to the student, but only global \textit{views} to the teacher. While this task is intrinsically difficult in the surgical domain, for the previously discussed reasons, it may be made even more challenging by the presence of the \textit{Strong-Color} augmentation, leading to unreliable feature representations.
\\
\begin{figure}[h!]
  \begin{subfigure}
    \centering
    \includegraphics[width=0.48\linewidth]{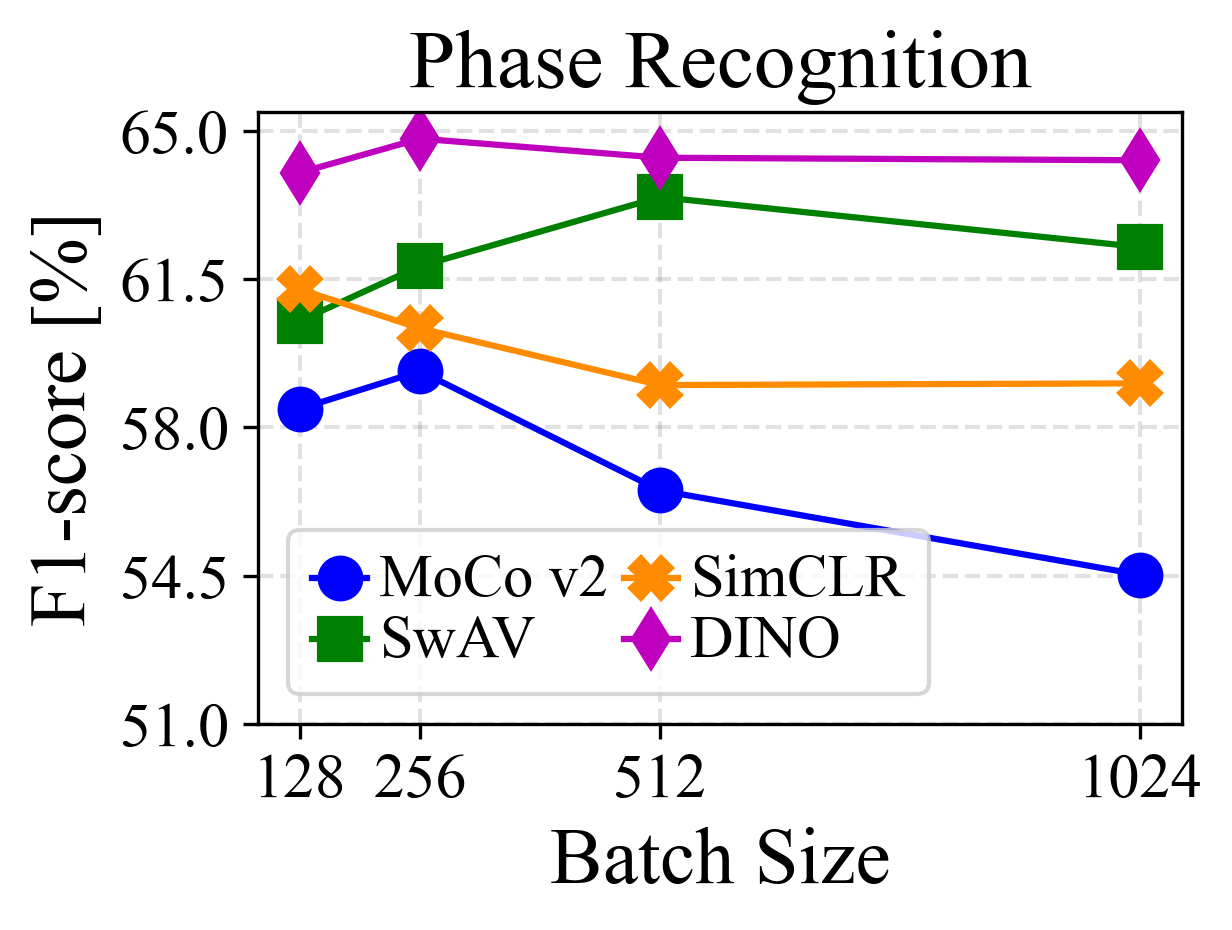}
  \end{subfigure}
  \begin{subfigure}
    \centering
    \includegraphics[width=0.48\linewidth]{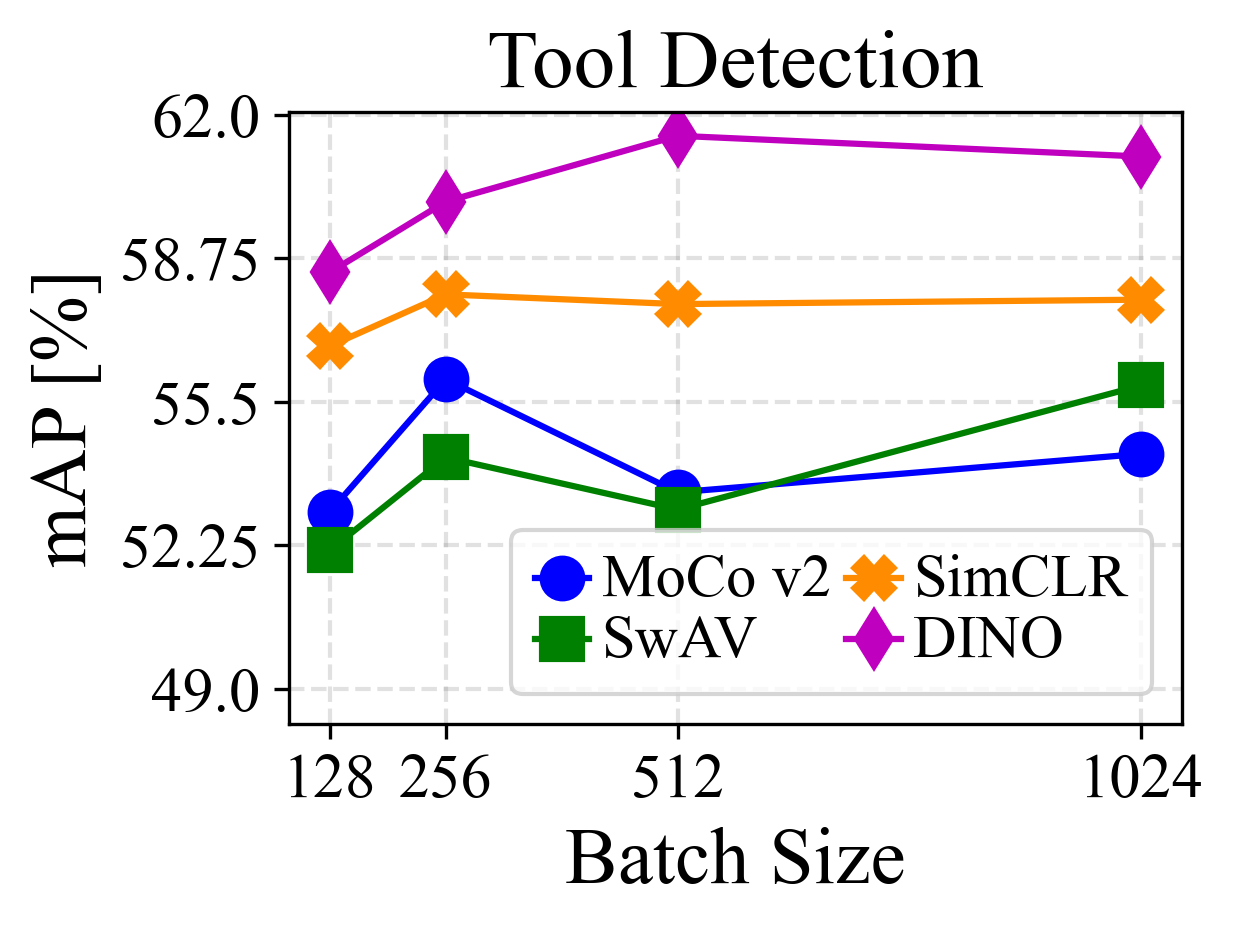}  
  \end{subfigure}
  \caption{Performance of each method on Cholec80 varying the batch size used for self-supervised pretraining. Results were obtained using linear evaluation on the validation set. Left: $F_{1}$-score for phase recognition. Right: mAP for tool presence detection.}
  \label{fig:abl_batch}
\end{figure}

\noindent\textbf{Batch size. } Overall, larger batch sizes do not improve feature quality. Clear improvements are only perceivable between 128 and 256 (up to 4.8\% $F_{1}$ for phase recognition, 5.6\% mAP for tool detection) across all tasks and methods - except for phase recognition with SimCLR. Results for 256 and above, however, generally contradict claims from other SSL works \citep{chen2020simple, caron2020unsupervised, caron2021emerging}, especially on the phase recognition task (Fig. \ref{fig:abl_batch}): from 256 to 1024, MoCo v2's $F_{1}$ score drops by 5.5\%. No clear positive impact of increasing batch size past 256 can be seen on tool presence detection either (Fig. \ref{fig:abl_batch}).

This inconsistency with results obtained on natural images is possibly due to differences in data scale since Cholec80 (at 1 fps: $\sim 10^5$ samples, $7$ classes) is far smaller than ImageNet ($>10^6$ samples, $10^3$ classes). During training, batches are therefore sampled under completely different conditions; since SSL methods, in the absence of labels, rely heavily on negative and positive samples to separate classes, this can affect the final performance. 

In the literature, one documented adverse effect of larger batches in SSL is shown by \cite{chen2020simple} on SimCLR, when the batch size is pushed up to high values ($>$2048). A scaled-back version of this phenomenon might be at play here.
\\
\begin{figure}[h!]
  \begin{subfigure}
    \centering
    \includegraphics[width=0.48\linewidth]{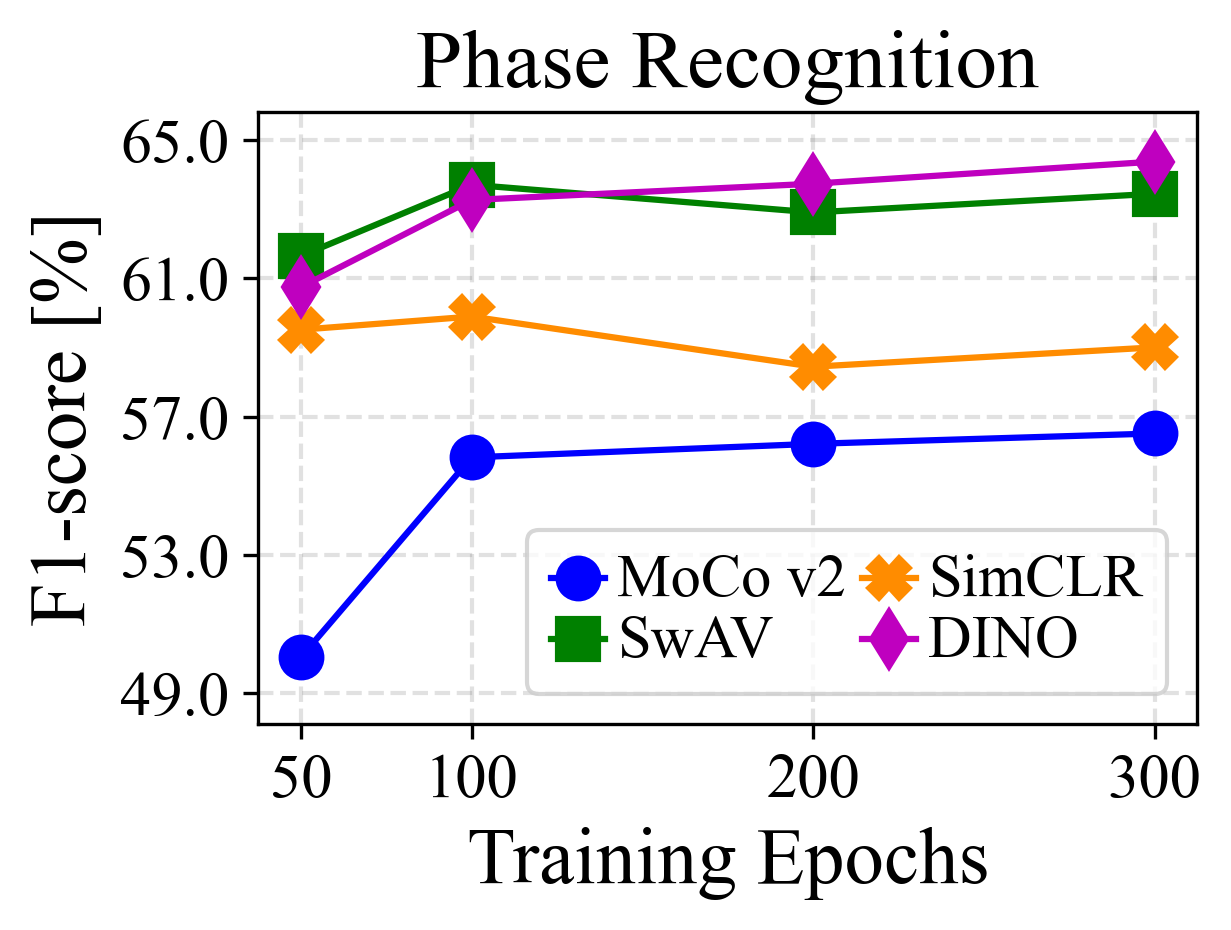}
  \end{subfigure}
  \begin{subfigure}
    \centering
    \includegraphics[width=0.48\linewidth]{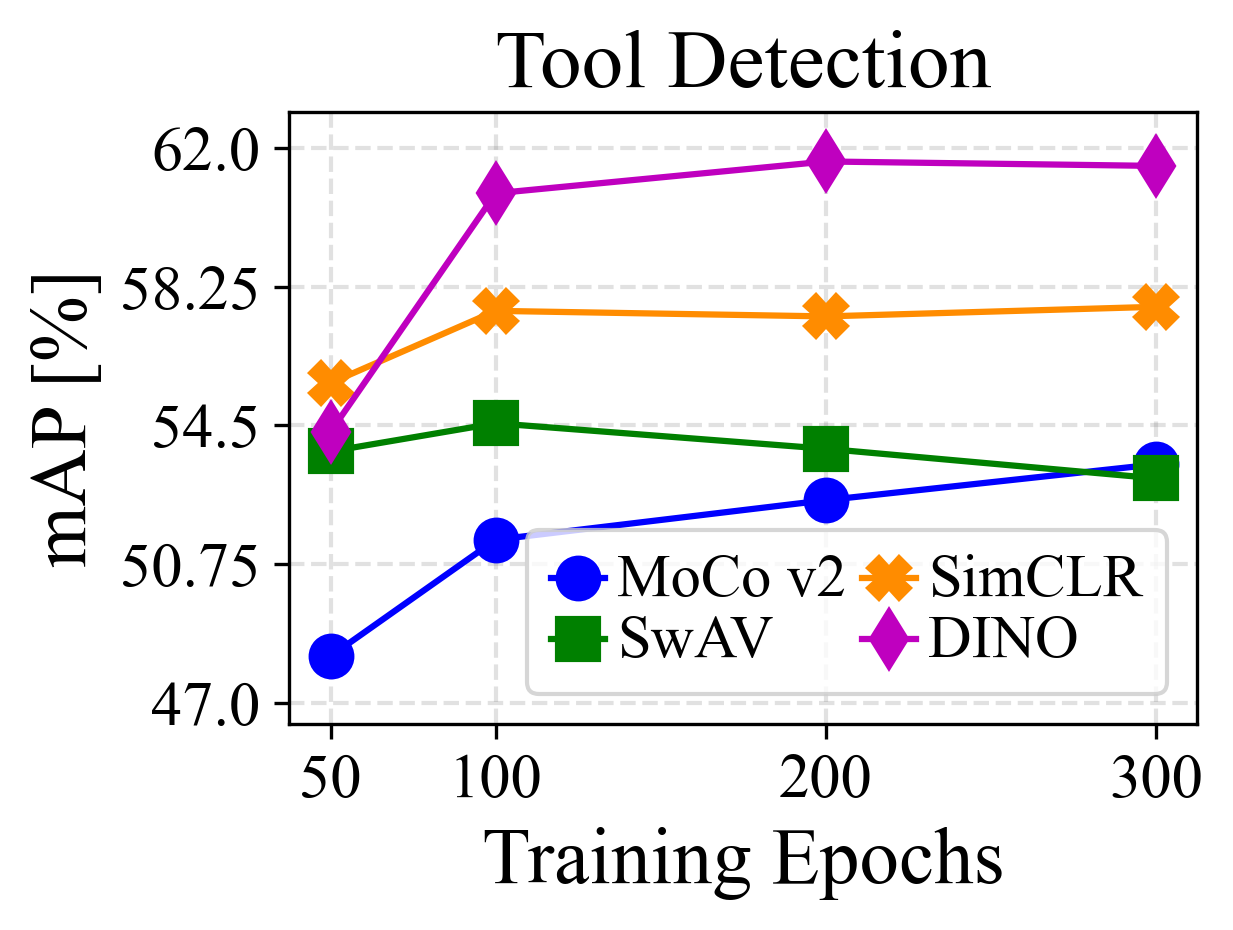} 
  \end{subfigure}
  \caption{Performance of each method on Cholec80 varying the number of epochs used for self-supervised pretraining. Results were obtained using linear evaluation on the validation set. Left: $F_{1}$-score for phase recognition. Right: mAP for tool presence detection.}
  \label{fig:abl_epoch}
\end{figure}

\noindent\textbf{Epochs. } Overall, phase recognition and tool presence detection performance (Fig. \ref{fig:abl_epoch}) tends to saturate as epochs increase, with nuances from one SSL method to another. SwAV and SimCLR in particular clearly peak earlier than the other two methods at 100 epochs, losing up to 2\% phase recognition $F_{1}$ and 2\% tool presence detection mAP afterward.
In contrast, MoCo v2 and DINO improve over the entire 300-epoch training period, with, nonetheless, a noticeable slowdown after 100 epochs.

This disparity could be a result of including a momentum encoder (used by both MoCo v2 and DINO). 
The momentum encoder enables a greater diversity in pairs of latent vectors generated by the network backbone during training: in MoCo v2, via a greater set of negative samples to choose from, and in DINO, via the teacher network incorporating context from a wider variety of samples.
Consequently, longer training may allow models to learn more robust representations.
\\
\begin{figure}[h!]
  \begin{subfigure}
    \centering
    \includegraphics[width=0.48\linewidth]{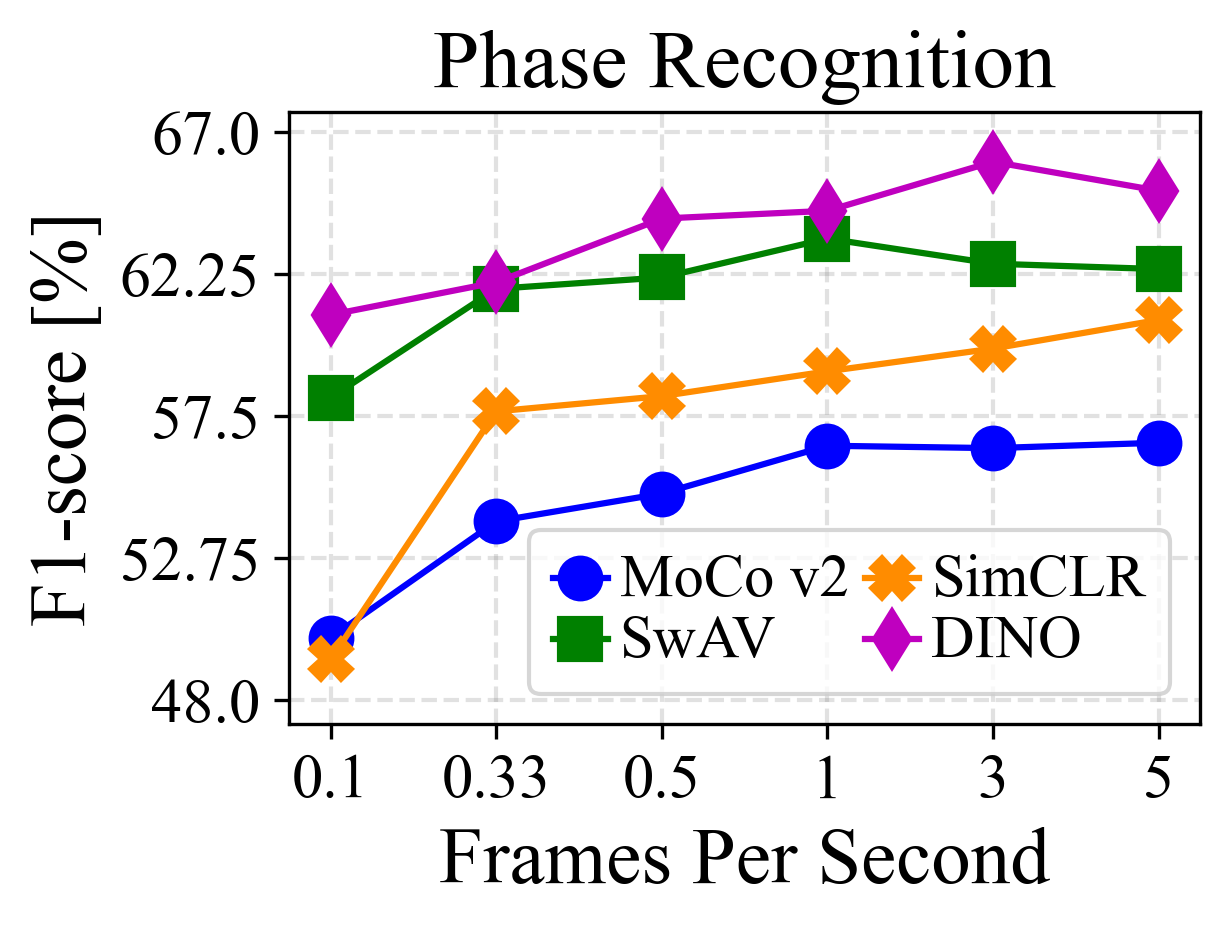}
  \end{subfigure}
  \begin{subfigure}
    \centering
    \includegraphics[width=0.48\linewidth]{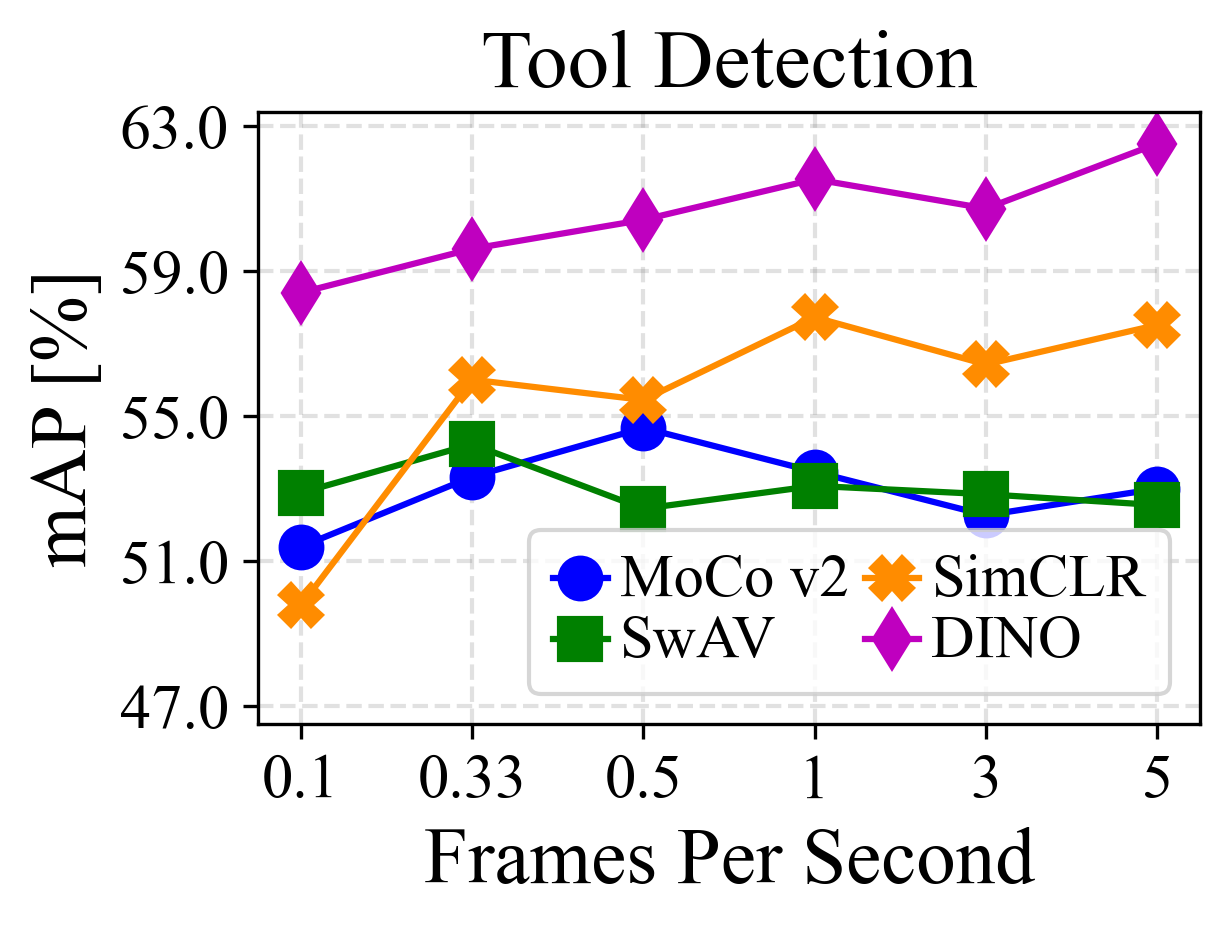} 
  \end{subfigure}
  \caption{Performance of each method on Cholec80 varying the Frames Per Second for self-supervised pretraining. Results were obtained using linear evaluation on the validation set. Left: $F_{1}$-score for phase recognition. Right: mAP for tool presence detection.}
  \label{fig:abl_fps}
\end{figure}

\begin{table*}[] 
\centering
\small
\caption{The average results across methods are presented for phase recognition, tool presence detection and the average across both tasks (Selection metric). For each individual ablation, results are presented in descending order of performance according to the Selection metric. The Setting column refers to the value of the parameter being ablated, while all other settings are kept to the default values specified in Table \ref{tab:hp_defaults}. For the augmentation ablation, we use the following notations: MC - Multi-Crop, C - Color, G - Geometric, S - Strong-color; for the MC setting columns, we specify the total number of crops used (including 2 high-resolution crops) and for the S, G, and C setting columns, we specify whether those augmentation categories were included or ``on''.}
\label{tab:selection_criteria}
\begin{tabular*}{\textwidth}{llp{1.7cm}ll||p{2cm}p{0.25cm}p{0.25cm}p{0.25cm}p{0.4cm}p{1.7cm}p{1cm}l}\hline
\multirow{2}{*}{\textbf{Ablation}}       & \multirow{2}{*}{\textbf{\textbf{Setting}}} & \textbf{Selection} & Phase & Tool & \multirow{2}{*}{\textbf{Ablation}}             & \multicolumn{4}{c}{\textbf{Setting}}                                                            & \textbf{Selection} & Phase & Tool \\\cline{7-10}
                                &                          &        \textbf{metric}                           &              (F1)                &       (mAP)                       &                                       & MC & C                         & G                         & S                         &   \textbf{metric}                                &   (F1)                          &       (mAP)                       \\\hline
\multirow{6}{*}{\textbf{Sampling rate}}  & 5.0                      & 58.8                              & 61.2                        & 56.4                        & \multirow{17}{*}{\textbf{Augmentations}} & 2  & \checkmark & \checkmark & \xmark     & 60.0                              & 63.5                        & 56.5                        \\
                                & 1.0                      & 58.6                              & 60.8                        & 56.4                        &                                       & 2  & \checkmark & \checkmark & \checkmark & 59.6                              & 63.2                        & 55.9                        \\
                                & 3.0                      & 58.4                              & 61.2                        & 55.6                        &                                       & 4  & \checkmark & \checkmark & \checkmark & 59.1                              & 61.7                        & 56.5                        \\
                                & 0.5                      & 57.8                              & 59.8                        & 55.7                        &                                       & 4  & \checkmark & \checkmark & \xmark     & 58.9                              & 61.1                        & 56.8                        \\
                                & 0.33                     & 57.3                              & 58.8                        & 55.8                        &                                       & 8  & \checkmark & \checkmark & \xmark     & 58.6                              & 60.8                        & 56.4                        \\
                                & 0.1                      & 53.9                              & 54.6                        & 53.1                        &                                       & 8  & \checkmark & \checkmark & \checkmark & 54.7                              & 56.4                        & 53.1                        \\\cline{1-5}
\multirow{4}{*}{\textbf{Batch size}}     & 256                      & 59.3                              & 61.6                        & 57.1                        &                                       & 2  & \xmark     & \checkmark & \xmark     & 53.7                              & 55.4                        & 52.0                        \\
                                & 1024                     & 58.6                              & 60.0                        & 57.3                        &                                       & 2 & \xmark & \checkmark & \checkmark                                                                  & 53.3                              & 54.6                        & 51.9                        \\
                                & 512                      & 58.6                              & 60.8                        & 56.4                        &   \multicolumn{7}{c}{\phantom{centering-space--add}\textbf{...}}         \\
                                & 128                      & 58.1                              & 61.1                        & 55.1                        &                                       & 8  & \checkmark & \xmark     & \checkmark & 45.5                              & 47.5                        & 43.6                        \\\cline{1-5}
\multirow{3}{*}{\textbf{Initialization}} & FS                 & 62.7                              & 64.4                        & 60.9                        &                                       & 4  & \xmark     & \xmark     & \checkmark & 41.2                              & 42.2                        & 40.2                        \\
                                & Rand                  & 58.6                              & 60.8                        & 56.4                        &                                       & 2  & \xmark     & \xmark     & \checkmark & 40.2                              & 41.1                        & 39.4                        \\
                                & SS             & 57.9                              & 58.9                        & 56.8                        &                                       & 8  & \xmark     & \xmark     & \checkmark & 37.3                              & 37.8                        & 36.8                        \\\cline{1-5}
\multirow{4}{*}{\textbf{Epochs}}         & 300                      & 58.6                              & 60.8                        & 56.4                        &                                       & 4  & \xmark     & \xmark     & \xmark     & 37.3                              & 36.9                        & 37.6                        \\
                                & 100                      & 58.4                              & 60.7                        & 56.1                        &                                       & 2  & \xmark     & \xmark     & \xmark     & 37.0                              & 37.2                        & 36.8                        \\
                                & 200                      & 58.3                              & 60.3                        & 56.4                        &                                       & 8  & \xmark     & \xmark     & \xmark     & 36.8                              & 35.8                        & 37.7                        \\
                                & 50                       & 55.5                              & 58.0                        & 53.0                        &                                       & 8  & \xmark     & \checkmark & \checkmark & 33.1                              & 31.4                        & 34.8  \\\hline

\end{tabular*}
\end{table*}
\noindent\textbf{Sampling rate. } As previously stated, surgical videos pose a particularly interesting technical setting for SSL research in general because surgical videos often provide a very stable context while the anatomy in the scene is manipulated. While increasing the number of frames sampled per second could dramatically expand the available training data, performance might not increase due to redundancy. Indeed, with the 5 sampling rates examined here, we observe marginal utility in sampling frames beyond a certain frequency. For both tasks, when sampling frames at over 1 fps, we observe no consistent improvement across methods or tasks when training with higher sampling rates (Fig. \ref{fig:abl_fps}). This is an important finding that may lend useful intuition to researchers applying SSL to domains with similar motion characteristics on how best to allocate computational resources, when training these intensive methods comes with a sizeable financial and environmental cost. To note, for a fair comparison, we perform this experiment here assuming an equal distribution of computational resources, i.e. we evaluated the models after performing self-supervised pretraining for the same number of iterations for each frame rate. This implies that the 1 fps experiments were trained for $\sim$ 5 times as many epochs as the 5 fps experiments.
\\
\begin{figure}[h!]
  \begin{subfigure}
    \centering
    \includegraphics[width=0.48\linewidth]{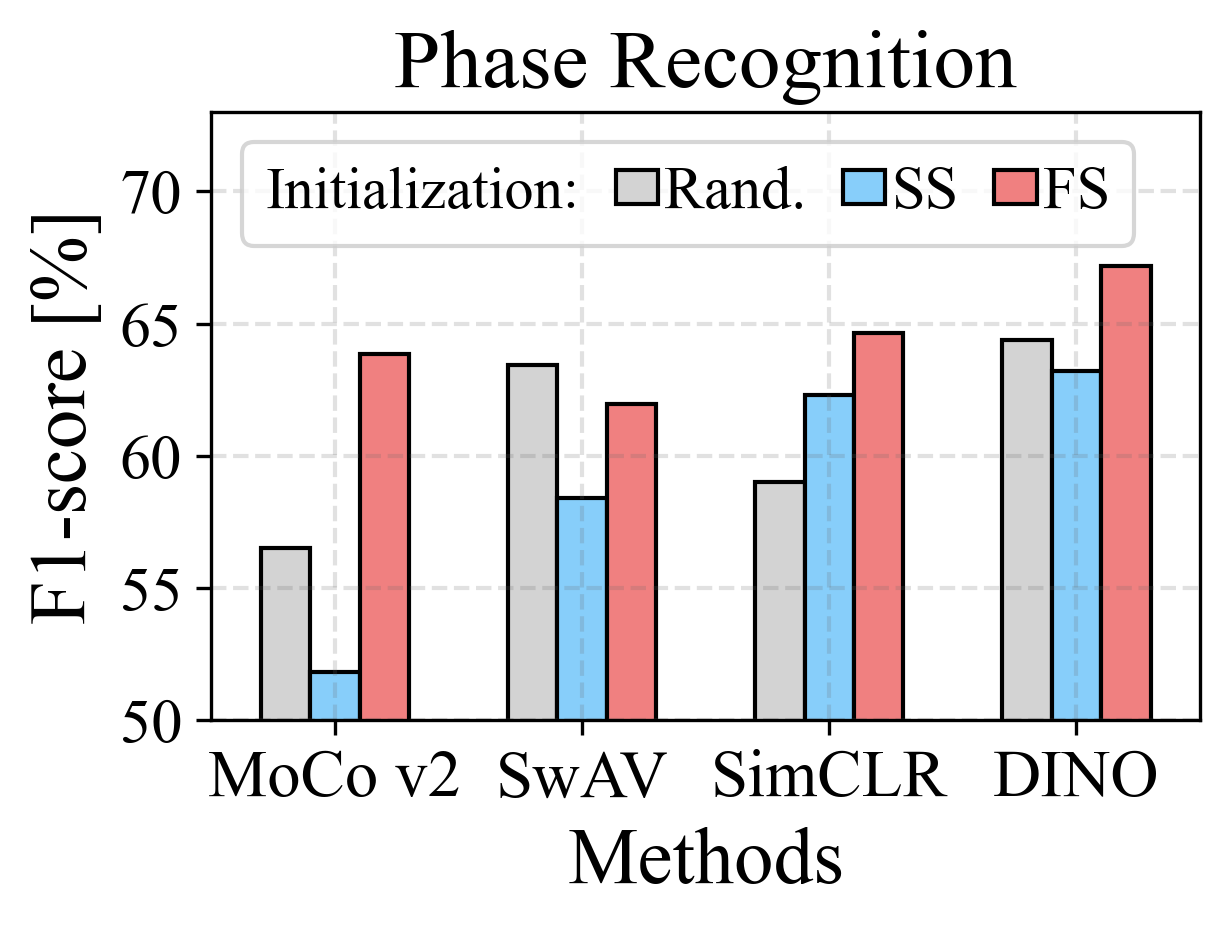}
  \end{subfigure}
  \begin{subfigure}
    \centering
    \includegraphics[width=0.48\linewidth]{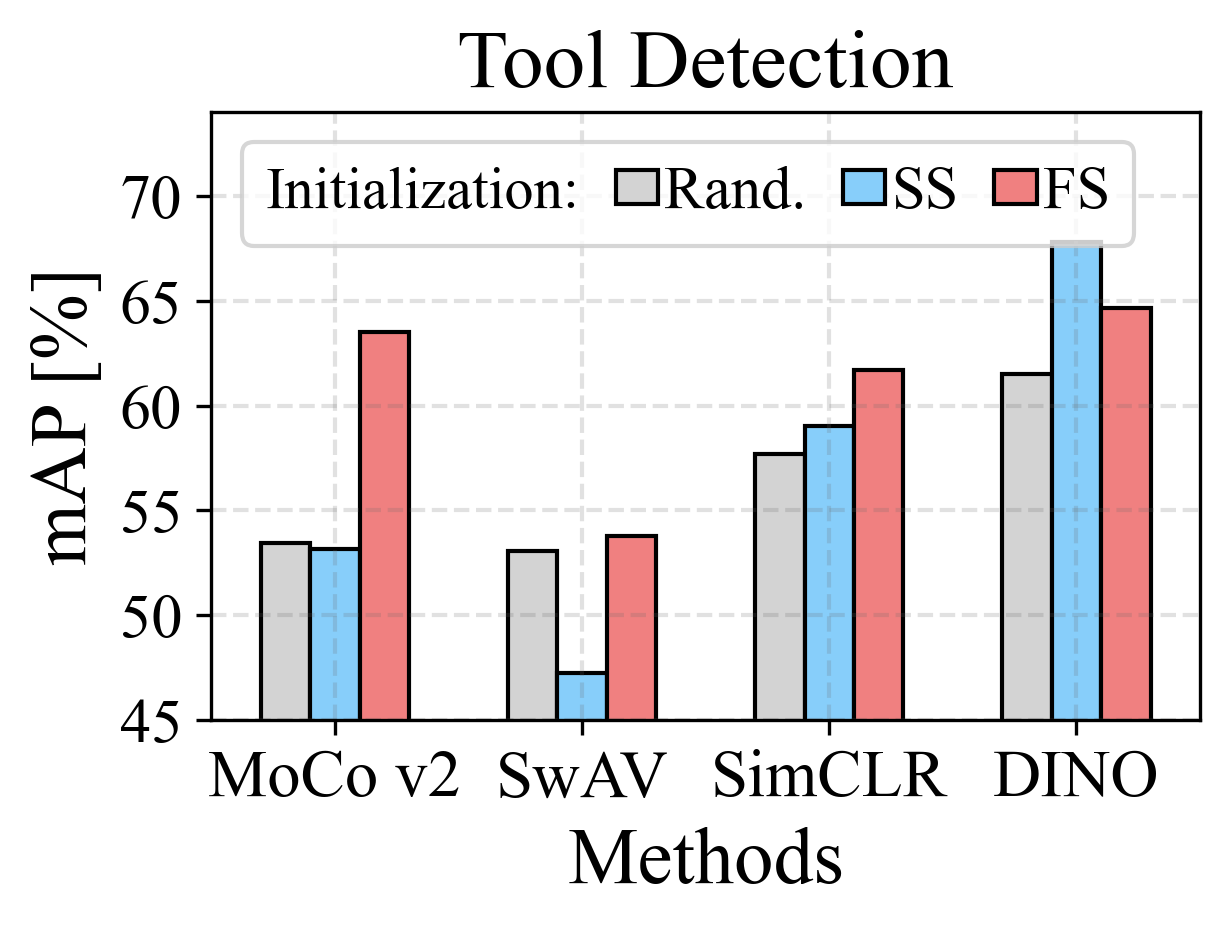} 
  \end{subfigure}
  \caption{Performance of each method on Cholec80 varying the network initialization strategies before performing self-supervised pretraining: random initialization (\textit{Rand.}), ImageNet self-supervised (\textit{SS}), ImageNet fully-supervised (\textit{FS}). Results were obtained using linear evaluation on the validation set. Left: $F_{1}$-score for phase recognition. Right: mAP for tool presence detection.}
  \label{fig:abl_init}
\end{figure}
\noindent\textbf{Initialization. } In general computer vision, the common practice for SSL experimentation is to train models to learn self-supervised representations entirely from scratch (i.e. random weights) before using these representations to attempt to replicate fully supervised performance - for image recognition on Imagenet, as a prominent example. Weights obtained in this manner are then intended to serve as initialization for downstream tasks. However, in surgical computer vision, Imagenet fully supervised weights are considered as a readily available resource: the practice of using them to initialize models is tacitly recognized as standard by the community. The choice of initialization is therefore not trivial, with 3 options available before starting SSL training on surgical data:
\begin{enumerate}
    \item ``Rand.'': randomly initializing weights
    \item ``SS'': initializing weights with self-supervised pretraining on ImageNet
    \item ``FS'': initializing weights with fully supervised pretraining on ImageNet
\end{enumerate}

Across all SSL methods (Fig. \ref{fig:abl_init}), models initialized with ``FS'' significantly outrank models with ``Rand.`` or ``SS.'' initialization; most noticeably with MoCo v2 (up to +12\% phase recognition $F_{1}$, +11\% tool detection mAP compared to the other two). Results between ``Rand.'' and ``SS.'' do not clearly favor one over the other. This is obviously a major difference from general computer vision, which expects models initialized from scratch to improve on any downstream task through SSL training. One explanation for this discrepancy could be the set of invariances learned in the natural domain, which may not apply to surgical images.

\noindent\textbf{Hyperparameter study conclusion. }
This study provides a detailed view of each SSL method's reaction to changes in parametrization when operating in the surgical domain, exposing noteworthy differences with the natural domain - regarding augmentations, batch size and initialization most prominently. However, when considering all four SSL methods and both tasks simultaneously, global trends can be difficult to clearly point out. To achieve this in a quantitative and principled manner, we define a selection metric, defined as the average of all phase recognition $F_{1}$ scores and tool presence detection mAPs across all methods for a given setting. Using this, we are able to rank the values of a given hyperparameter by overall performance across downstream tasks, and then retain the best. This forms a global set of \textbf{recommended settings} (Table \ref{tab:hp_defaults}) for SSL in the surgical domain.

In Table \ref{tab:selection_criteria}, we present the results ranked according to this selection metric for each ablation to facilitate the analysis of invariant trends for methods and tasks. For each hyperparameter, we summarize the trends in brief below:
\begin{itemize}
  \item Sampling rate: We observe only a marginal utility of increasing the sampling rate beyond a certain point, with the selection metric saturating past 0.33 fps.
  \item Batch size: The results show that for the considered tasks and dataset, SSL method performance is mostly robust to variations of batch size. Varying the batch size between 128-1024 results in a maximum variation of 1.1\% F1 and 2\% mAP on average across methods for phase recognition and tool presence detection, respectively. 
  \item Initialization: Initialization before self-supervised representation learning proves to be a critical hyperparameter with significant and consistent gains in performance across both methods and tasks. Initializing with Imagenet fully supervised (``FS'') weights proves to be the optimal setting amongst the considered initializations.
  \item Epochs: For both considered tasks, we see significant gains in performance up to 100 epochs after which it plateaus, with an average variation of 0.4\% F1 and 0.4\% mAP between 100 and 300 epochs.
  \item Augmentations: Interestingly, we observe largely consistent trends for different augmentation settings for both tasks. Color and geometric augmentations feature consistently in top-performing augmentation settings. On average across methods, the addition of multiple low-resolution views and strong color augmentations has a less clear impact on performance. 
\end{itemize}

\subsection{Data supply study}
The recommended choice of hyperparameters mentioned above provides, on average, close to optimal conditions for observing our panel of SSL methods in practical use cases, with varying quantities of labeled or unlabeled data. Our proposed usage of SSL is defined as follows: self-supervised training is performed in the surgical domain before finetuning for surgical downstream tasks.

\noindent\textbf{Labeled data supply. }
In this section of the data supply study, self-supervised training is first performed on the entire training set of Cholec80 with the recommended hyperparameters. Surgical downstream task finetuning is then applied using variable amounts of labeled data: 40 videos (100\% of the training set), or in semi-supervision with 10 videos (25\%) or 5 videos (12.5\%); for these last two settings, the portions of the training set are drawn following a stratified random sampling approach (see Sec. \ref{sec:metrics_and_splits}). Results for these experiments are reported in Tables \ref{tab:fcn_main} (phase recognition on single frames), \ref{tab:tcn_main} (phase recognition on videos with a temporal model), and \ref{tab:tool_main} (tool presence detection). We compare our proposed usage of SSL (``ours'') on Cholec80 using the recommended hyperparameters (Table \ref{tab:hp_defaults}) with the mode of operation borrowed from general computer vision (``base'') - i.e. finetuning directly from weights pretrained with SSL on Imagenet. The bottom row in each table (``No SSL'') provides an additional point of comparison, where we finetune models initialized with fully supervised Imagenet weights without any SSL.

\begin{table*}[th!]
  \centering
  \caption{{\color{changetext} Effect of our proposed SSL pretraining in the surgical domain (``Ours'') on surgical phase recognition performance from single frames. ``Base'' refers to self-supervised pretraining on Imagenet only. ``No SSL'' refers to fully supervised pretraining on Imagenet only. Bold indicates the best performance for a given number of labeled videos.}}
  \label{tab:fcn_main}
  \begin{tabular*}{\textwidth}{l @{\extracolsep{\fill}}cccccc}
  \hline
  \multicolumn{7}{c}{\textbf{Surgical phase recognition $F_{1}$ - single frame}}                      \\ \hline
  Labels & \multicolumn{2}{c}{40 videos} & \multicolumn{2}{c}{10 videos} & \multicolumn{2}{c}{5 videos} \\ \hline
         & Base    & \textbf{Ours}   & Base   & \textbf{Ours}   & Base    & \textbf{Ours}    \\
  \textbf{DINO}   & 71.6 & 71.1& 60.6 $\pm$ 0.6 & 62.2 $\pm$ 0.9 & 51.4 $\pm$ 5.1 & 56.3 $\pm$ 4.8\\
  \textbf{MoCo v2}   & 70.3 & 71.3& 58.5 $\pm$ 0.6 & \textbf{64.4 $\pm$ 1.7} & 52.1 $\pm$ 4.5 & \textbf{58.1 $\pm$ 5.3}\\
  \textbf{SimCLR}   & 70.3 & \textbf{71.8}& 58.9 $\pm$ 2.4 & 63.5 $\pm$ 1.1 & 51.3 $\pm$ 3.9 & 57.2 $\pm$ 5.0\\
  \textbf{SwAV} & 70.2 & 70.3& 58.8 $\pm$ 0.9 & 62.2 $\pm$ 1.9 & 50.9 $\pm$ 4.5 & 57.1 $\pm$ 3.7\\ \hline
  \textbf{No SSL} & \multicolumn{2}{c}{71.5}      & \multicolumn{2}{c}{60.4 $\pm$ 0.4}     & \multicolumn{2}{c}{52.0 $\pm$ 6.5} \\ \hline
  \end{tabular*}
\end{table*}
\begin{table*}[th!]
  \centering
  \caption{{\color{changetext} Effect of our proposed SSL pretraining in the surgical domain (``Ours'') on surgical phase recognition performance from videos when finetuning a temporal model (TCN - \cite{Czempiel2020TeCNOSP}) on top of the backbones described in Table \ref{tab:fcn_main}. Bold indicates the best performance for a given amount of labeled videos.}}
  \label{tab:tcn_main}
  \begin{tabular*}{\textwidth}{l @{\extracolsep{\fill}}cccccc}
  \hline
  \multicolumn{7}{c}{\textbf{Surgical phase recognition $F_{1}$ - temporal}}                      \\ \hline
  Labels & \multicolumn{2}{c}{40 videos} & \multicolumn{2}{c}{10 videos} & \multicolumn{2}{c}{5 videos} \\ \hline
         & Base    & \textbf{Ours}   & Base   & \textbf{Ours}   & Base    & \textbf{Ours}    \\
  \textbf{DINO}   & 81.5 & \textbf{81.6}	&	71.3 $\pm$ 0.6 & 70.4 $\pm$ 0.4 &	61.1 $\pm$ 9.0 & 65.0 $\pm$ 5.4 \\
  \textbf{MoCo v2}   & 79.5 & 79.6	&	69.1 $\pm$ 1.8 & \textbf{74.1 $\pm$ 0.4}	&	63.4 $\pm$ 4.3 & 66.1 $\pm$ 4.2 \\
  \textbf{SimCLR} & 78.8 & 81.1	&	69.2 $\pm$ 2.4 & 72.5 $\pm$ 0.4	&	63.6 $\pm$ 3.9 & 66.6 $\pm$ 2.4 \\
  \textbf{SwAV} & 78.4 & 79.5	&	68.7 $\pm$ 0.5 & 71.4 $\pm$ 0.7	&	60.9 $\pm$ 7.0 & \textbf{68.3 $\pm$ 1.3} \\
   \hline
  \textbf{No SSL} & \multicolumn{2}{c}{80.3}      & \multicolumn{2}{c}{70.1 $\pm$ 0.2}     & \multicolumn{2}{c}{62.3 $\pm$ 7.4} \\ \hline
  \end{tabular*}
\end{table*}
\begin{table*}[th!]
  \centering
  \caption{Effect of our proposed SSL pretraining in the surgical domain (``Ours'') on surgical tool presence detection performance. Bold indicates the best performance for a given amount of labeled videos.}
  \label{tab:tool_main}
  \begin{tabular*}{\textwidth}{l @{\extracolsep{\fill}}cccccc}
  \hline
  \multicolumn{7}{c}{\textbf{Surgical tool presence detection mAP}}                      \\ \hline
  Labels & \multicolumn{2}{c}{40 videos} & \multicolumn{2}{c}{10 videos} & \multicolumn{2}{c}{5 videos} \\ \hline
         & Base    & \textbf{Ours}   & Base   & \textbf{Ours}   & Base    & \textbf{Ours}    \\
  \textbf{DINO}   &92.1& 93.2& 70.1 $\pm$ 2.7 & 81.2 $\pm$ 1.4 & 50.6 $\pm$ 1.6 & 68.7 $\pm$ 2.3\\
  \textbf{MoCo v2}   &92.9& 93.5& 70.4 $\pm$ 1.3 & \textbf{85.7 $\pm$ 1.1} & 56.5 $\pm$ 3.3 & \textbf{74.7 $\pm$ 1.8}\\
  \textbf{SimCLR} &90.4& 93.1& 66.7 $\pm$ 0.1 & 83.0 $\pm$ 0.9 & 49.3 $\pm$ 1.4 & 69.7 $\pm$ 3.0\\
  \textbf{SwAV}   &92.5& 92.8& 70.5 $\pm$ 1.5 & 79.1 $\pm$ 1.7 & 52.5 $\pm$ 1.8 & 63.0 $\pm$ 0.7\\ \hline
  \textbf{No SSL} & \multicolumn{2}{c}{\textbf{93.6}}      & \multicolumn{2}{c}{77.9 $\pm$ 0.8}     & \multicolumn{2}{c}{60.0 $\pm$ 2.3}\\ \hline
  \end{tabular*}
\end{table*}

{\color{changetext} In most low-label settings (10, 5 videos), adding any of the 4 SSL methods systematically improves performance on both surgical tasks, compared to direct finetuning from supervised Imagenet weights without SSL. This improvement reaches up to 6.1\% (5 videos, MoCo v2) for single-frame phase recognition, 6\% (5 videos, SwAV) for temporal phase recognition, and 14.7\% for tool presence detection (5 videos, MoCo v2). Gains are consistently observed, especially in low-label settings where standard deviation across splits mostly stays underneath 3\% (32 out of 48 table entries). 100\% label availability tends to saturate performance on downstream tasks, leaving little room for improvement from SSL; still, results are on par with those obtained without SSL for both tool presence detection (mostly $<1\%$ difference) and phase recognition, with the largest deficit (-1.2\%) recorded for SwAV on single frames. Out of the four SSL methods presented here, MoCo v2 seems to yield better results, 5 times achieving the best performance for a given number of labeled videos.

Most importantly, these results challenge the generalizability of general computer vision SSL. As demonstrated in \cite{oord2018representation, he2020momentum, chen2020improved, caron2021emerging}, self-supervised pretraining on natural images enhances downstream task performance in the natural image domain; however, these gains may not carry over to more complex and more specific domains. Indeed, when pretrained on Imagenet, rarely do any of the SSL methods featured here improve performance on surgical downstream tasks, compared to the ``No SSL'' baseline (only 7 out of 36 times). For phase recognition, this usage of SSL can cause $F_{1}$ score to drop by up to 1.9\%, while for tool presence detection, the degradation reaches up to 11.2\% mAP. Overall, our proposed use of SSL outperforms the ``base'' usage by up to 6.2\% on single-frame phase recognition, 7.4\% on temporal phase recognition, and 20.4\% on tool presence detection.}

\begin{table*}[th!]
  \centering
  \caption{{\color{changetext}External comparison with \cite{shi_surgssl} for semi-supervised surgical phase recognition. Bold indicates the best performance for a given amount of labeled videos used for finetuning.}}
  \label{tab:cuhk}
  \begin{tabular*}{\textwidth}{l @{\extracolsep{\fill}}llccc}
  \hline
  \multicolumn{6}{c}{\textbf{External comparison - surgical phase recognition $F_{1}$}}                   \\ \hline
  \multicolumn{3}{l}{Labels}                                                      & 40 videos & 10 videos  & 5 videos\\ \hline
  \textbf{External}         & \multicolumn{2}{l}{\textbf{NL-RCNet}}           & 82.1  & 73.5 & 67.3     \\
                  {\small \textit{quoted from \cite{shi_surgssl}}}              & \multicolumn{2}{l}{\textbf{NL-RCNet+}}          & 84.4  & -     & -  \\
                                & \multicolumn{2}{l}{\textbf{CNN-BiLSTM-CRF}}     & -     & 75.3  & 70.9  \\
                                & \multicolumn{2}{l}{\textbf{MT}}                 & -     & 77.3  & 71.0  \\
                                & \multicolumn{2}{l}{\textbf{SurgSSL}}            & -     & 80.6  & 78.6  \\ \hline
  \textbf{Selected SSL methods} & \textbf{DINO}                    & single frame & 77.6   & 64.0  & 65.4  \\
    {\small \textit{metric and split from \cite{shi_surgssl}}}&    & temporal     & 91.8   & 81.1  & 76.9  \\
                                & \textbf{MoCo v2}                 & single frame & 81.7   & 72.6  & 69.3  \\
                                &                                  & temporal     & 91.3   & 82.5  & \textbf{81.4}  \\
                                & \textbf{SimCLR}                  & single frame & 84.5   & 73.8  & 67.0  \\
                                &                                  & temporal     &\textbf{93.6}  & \textbf{85.0}  & 80.0  \\
                                & \textbf{SwAV}                    & single frame & 86.1   & 67.1  & 69.5  \\
                                &                                  & temporal     & 91.0   & 79.8  & 80.7  \\ \hline
  \textbf{Baselines}            & \textbf{No SSL}                  & single frame & 81.0   & 65.6  & 60.8  \\
            {\small \textit{metric and split from \cite{shi_surgssl}}}& & temporal&87.4   & 81.5  & 78.4  \\ \hline
  \end{tabular*}
\end{table*}

{\color{changetext} Finally, we add an external comparison in Table \ref{tab:cuhk} with preexisting semi-supervised studies in surgical computer vision, based on results presented by \cite{shi_surgssl} for semi-supervised phase recognition on Cholec80, and using the same split and metric definition. As expected, selected SSL methods applied to single-frame models are often outranked by other approaches, by up to 16.6\% (DINO vs SurgSSL, 10 videos); however the external methods, we compare against, use temporal modeling, which gives them a strong advantage. For a fairer comparison, we examine models trained with our selected SSL methods used in conjunction with a temporal model (TCN): in these situations, they surpass preexisting semi-supervised approaches by a substantial amount - up to 14.1\%. Top $F_{1}$ scores are achieved by SimCLR (93.6\%, labels on 40 videos - 85.0\%, labels on 10 videos) and MoCo v2 (81.4\%, labels on 5 videos). To note, the architecture we use is fairly simple (CNN - TCN) compared to the more refined designs featured in the external methods; therefore our performance gains derive from the SSL methodology itself, and could further increase with more advanced architectures. These observations strongly confirm the high value of bringing SSL innovations from general computer vision to the surgical domain.}

\begin{figure}[ht!]
    \includegraphics[width=1.0\linewidth]{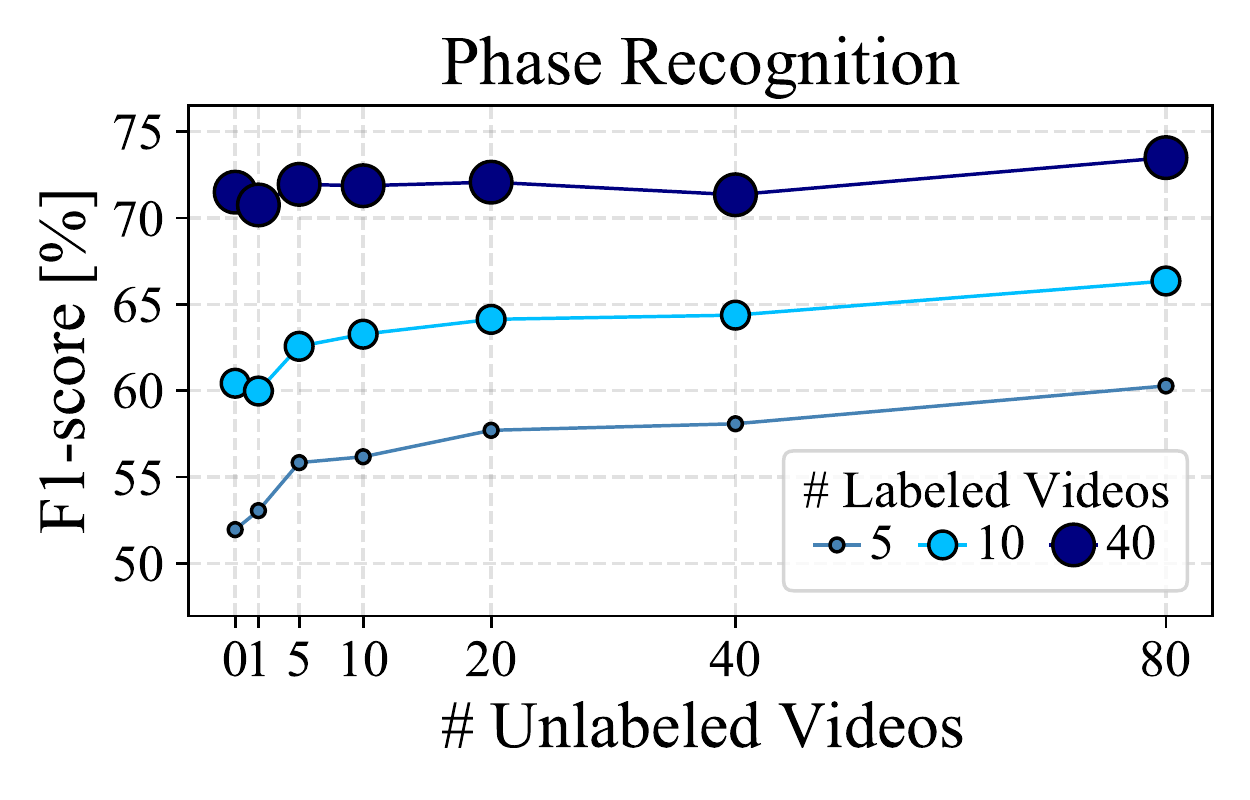}
    \caption{Single-frame phase recognition performance of MoCo v2 w.r.t. the number of unlabeled videos used for self-supervised pretraining, with finetuning on 5, 10, and 40 labeled videos.}
    \label{fig:data_supply_phase}
\end{figure}

\begin{figure}[ht!]
    \includegraphics[width=1.0\linewidth]{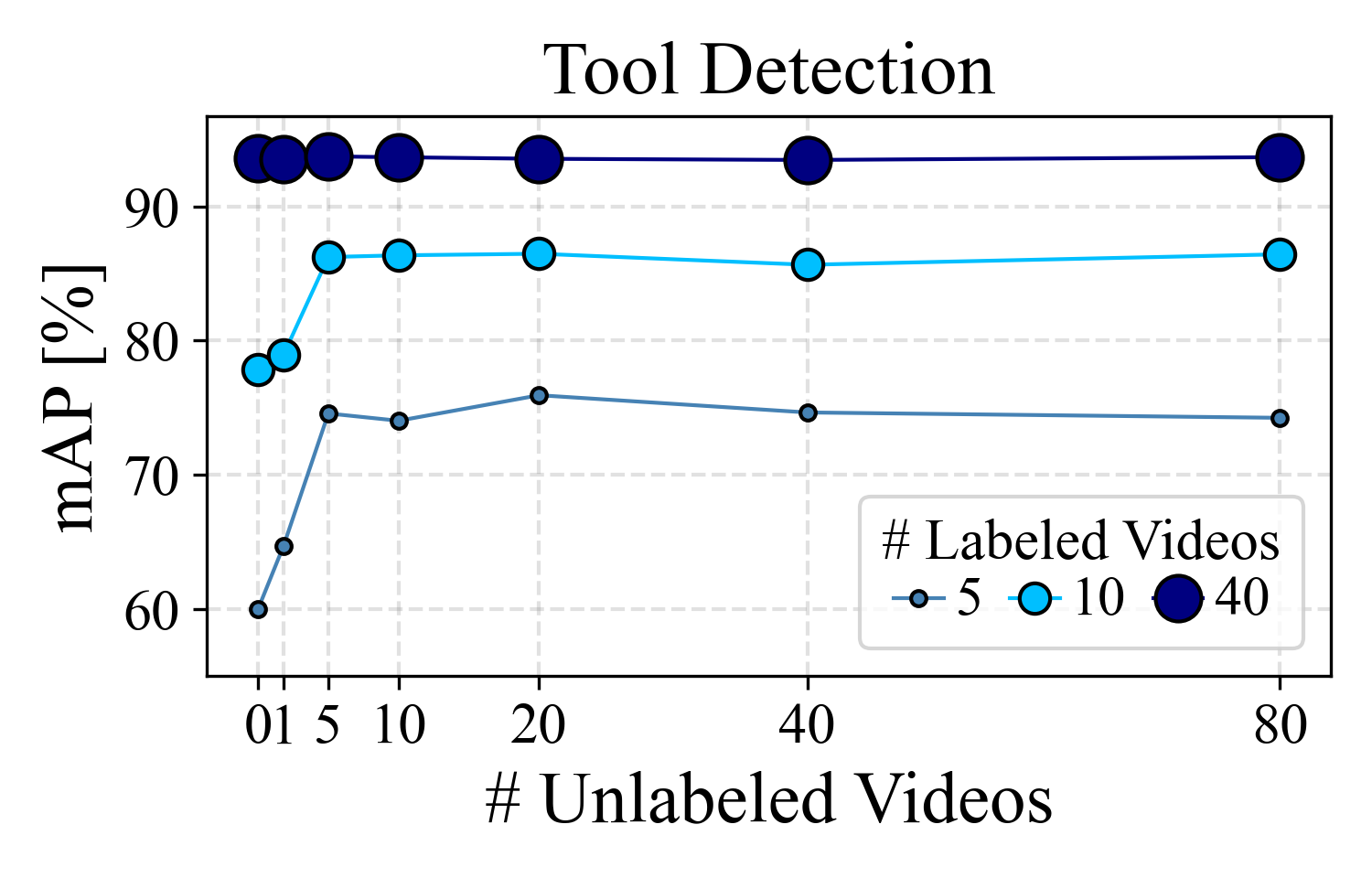}
    \caption{Tool presence detection performance of MoCo v2 w.r.t. the number of unlabeled videos used for self-supervised pretraining, with finetuning on 5, 10, and 40 labeled videos.}
    \label{fig:data_supply_tools}
\end{figure}

\noindent\textbf{Unlabeled data supply. } Our main experiments examined the performance of SSL in the surgical domain with a fixed quantity of unlabeled data for self-supervised pretraining; in this complementary set of experiments, we observe how SSL reacts when the quantity of unlabeled videos varies. This part of the study is conducted with MoCo v2 exclusively. Overall, our results (Fig. \ref{fig:data_supply_phase} and \ref{fig:data_supply_tools}) confirm a valuable benefit of SSL: for the most part, expanding unlabeled data - which is far easier than generating additional annotations - leads to increased performance in downstream surgical tasks. Particularly when few labeled instances are available, we see extremely pronounced improvements brought about by introducing SSL. For example, when only 5 labeled videos are available, self-supervised pretraining on just 10 unlabeled videos adds {\color{changetext}4.2\%} $F_{1}$ for phase recognition and {\color{changetext}$\sim$14.2\%} mAP for tool presence detection. These results further reinforce the practicality of utilizing these SSL methods in surgical applications, where working with small datasets is often the norm rather than the exception. We observe, however, two main limitations.

\begin{table*}[]
\centering
\begin{tabular}{llllcccccc}
\hline
\multicolumn{10}{c}{\textbf{Generalization Experiments}}\\\hline
\multicolumn{1}{c}{} & \multicolumn{3}{l}{\textbf{Dataset - architecture}} & \multicolumn{2}{c}{Labeled videos}    & \multicolumn{2}{c}{Labeled videos}        & \multicolumn{2}{c}{Labeled videos}       \\
\multicolumn{1}{c}{\multirow{-2}{*}{\textbf{Exp \#}}} & \cellcolor[HTML]{C0C0C0}SSL Dataset & \cellcolor[HTML]{00D2CB}Task         & \multicolumn{1}{r}{\cellcolor[HTML]{FFCCC9}Metric} & No SSL               & \textbf{MoCo v2}     & No SSL               & \textbf{MoCo v2}        & No SSL               & \textbf{MoCo v2}        \\\hline
& \multicolumn{3}{l}{\textbf{HeiChole - TCN head}} & \multicolumn{2}{c}{24 videos} & \multicolumn{2}{c}{4 videos}    & \multicolumn{2}{c}{2 videos}\\
                                                      
1                                                     & \cellcolor[HTML]{C0C0C0}Cholec80    & \cellcolor[HTML]{00D2CB}Phase        & \cellcolor[HTML]{FFCCC9}$F_{1}$                        & 58.6                 & \textbf{64.7}        & 41.7 $\pm$ 4.7       & \textbf{51.1 $\pm$ 3.3}          & 27.6 $\pm$ 6.0       & \textbf{39.0 $\pm$ 1.2}          \\
                                                      & \multicolumn{3}{l}{\textbf{HeiChole - linear head}}                                                                             &                      &                      &                      &                         &                      &                         \\
2                                                     & \cellcolor[HTML]{C0C0C0}Cholec80    & \cellcolor[HTML]{00D2CB}Tool         & \cellcolor[HTML]{FFCCC9}mAP                        & 62.5                 & \textbf{66.9}        & 36.7 $\pm$ 2.9       & \textbf{43.7 $\pm$ 0.4} & 25.1 $\pm$ 6.1       & \textbf{30.3 $\pm$ 2.3}          \\\hline
& \multicolumn{3}{l}{\textbf{CATARACTS - TCN head}} & \multicolumn{2}{c}{25 videos} & \multicolumn{2}{c}{6 videos}    & \multicolumn{2}{c}{3 videos}\\
3                                                     & \cellcolor[HTML]{C0C0C0}Cholec80    & \cellcolor[HTML]{00D2CB}Phase        & \cellcolor[HTML]{FFCCC9}$F_{1}$                       & \textbf{75.2}                 & 74.5                  & \textbf{65.7 $\pm$ 5.5}                  & 65.0 $\pm$ 5.6                     & \textbf{52.8 $\pm$ 4.7}                  &    50.7 $\pm$ 1.0                  \\
4                                                     & \cellcolor[HTML]{C0C0C0}CATARACTS   & \cellcolor[HTML]{00D2CB}Phase        & \cellcolor[HTML]{FFCCC9}$F_{1}$                       & 75.2                 & \textbf{77.2}                 & 65.7 $\pm$ 5.5       & \textbf{66.5 $\pm$ 3.8} & 52.8 $\pm$ 4.7       & \textbf{56.2 $\pm$ 5.5} \\
                                                      & \multicolumn{3}{l}{\textbf{CATARACTS - linear head}}                                                                            &                      &                      &                      &                         &                      &                         \\
5                                                     & \cellcolor[HTML]{C0C0C0}Cholec80    & \cellcolor[HTML]{00D2CB}Tool         & \cellcolor[HTML]{FFCCC9}mAP                        & \textbf{56.1}                 & 47.6                  & \textbf{37.7 $\pm$ 1.4}                  & 29.2 $\pm$ 2.2                     & \textbf{26.9 $\pm$ 1.6}                  & 19.0 $\pm$ 0.4                     \\
6                                                     & \cellcolor[HTML]{C0C0C0}CATARACTS   & \cellcolor[HTML]{00D2CB}Tool         & \cellcolor[HTML]{FFCCC9}mAP                        & 56.1                 & \textbf{57.3}        & 37.7 $\pm$ 1.4       & \textbf{40.8 $\pm$ 0.5} & 26.9 $\pm$ 1.6       & \textbf{31.2 $\pm$ 4.2} \\\hline
& \multicolumn{3}{l}{\textbf{CholecT50 - linear head}}  & \multicolumn{2}{c}{40 videos} & \multicolumn{2}{c}{10 videos}    & \multicolumn{2}{c}{5 videos}\\
7                                                     & \cellcolor[HTML]{C0C0C0}Cholec80    & \cellcolor[HTML]{00D2CB}Action       & \cellcolor[HTML]{FFCCC9}mAP                        & 19.4                 & \textbf{26.7}        & 14.4 $\pm$ 0.2       & \textbf{20.7 $\pm$ 0.2} & 11.2 $\pm$ 1.4       & \textbf{15.9 $\pm$ 0.8} \\
                                                      & \multicolumn{3}{l}{\textbf{CholecT50 - RDV head}}                                                                           &                      & \textbf{}            &                      & \textbf{}               &                      & \textbf{}               \\
8                                                    & \cellcolor[HTML]{C0C0C0}Cholec80    & \cellcolor[HTML]{00D2CB}Action       & \cellcolor[HTML]{FFCCC9}mAP                        & 31.4                 & \textbf{35.7}                 & 22.3 $\pm$ 1.8       & \textbf{25.5 $\pm$ 0.8}          & 14.9 $\pm$ 0.9       & \textbf{18.3 $\pm$ 1.2}          \\\hline
& \multicolumn{3}{l}{\textbf{Endoscapes - DeepLabv3+ head}} & \multicolumn{2}{c}{120 videos} & \multicolumn{2}{c}{30 videos}    & \multicolumn{2}{c}{15 videos}   \\
9                                                    & \cellcolor[HTML]{C0C0C0}Cholec80    & \cellcolor[HTML]{00D2CB}Segmentation & \cellcolor[HTML]{FFCCC9}$F_{1}$                       & \textbf{73.2}                 & \textbf{73.2}                 & 63.6 $\pm$ 1.0       & \textbf{64.3 $\pm$ 1.0}          & 58.1 $\pm$ 1.2       & \textbf{59.3 $\pm$ 1.7}          \\\hline
& \multicolumn{3}{l}{\textbf{CaDIS 8 classes - DeepLabv3+ head}} &  \multicolumn{2}{c}{19 videos} & \multicolumn{2}{c}{4 videos}    & \multicolumn{2}{c}{2 videos} \\
10  & \cellcolor[HTML]{C0C0C0}Cholec80    & \cellcolor[HTML]{00D2CB}Segmentation & \cellcolor[HTML]{FFCCC9}$F_{1}$ & 86.9 & \textbf{87.1} & 79.6 $\pm$ 1.6 & \textbf{82.5 $\pm$ 1.2}  & 79.5 $\pm$ 1.6 & \textbf{81.4 $\pm$ 1.2} \\
11  & \cellcolor[HTML]{C0C0C0}CaDIS & \cellcolor[HTML]{00D2CB}Segmentation & \cellcolor[HTML]{FFCCC9}$F_{1}$ & \textbf{86.9}& \textbf{86.9} & 79.6 $\pm$ 1.6 & \textbf{83.2 $\pm$ 0.8} & 79.5 $\pm$ 1.6 & \textbf{81.3 $\pm$ 0.8}\\                                               

& \multicolumn{3}{l}{\textbf{CaDIS 25 classes - DeepLabv3+ head}}&&&&&&\\
12 & \cellcolor[HTML]{C0C0C0}Cholec80    & \cellcolor[HTML]{00D2CB}Segmentation & \cellcolor[HTML]{FFCCC9}$F_{1}$ & \textbf{71.8} & 70.5 & 61.2 $\pm$ 1.9 & \textbf{62.4 $\pm$ 2.9} & 55.5 $\pm$ 5.8                  & \textbf{57.3 $\pm$ 6.7}                     \\
13                                                    & \cellcolor[HTML]{C0C0C0}CaDIS       & \cellcolor[HTML]{00D2CB}Segmentation & \cellcolor[HTML]{FFCCC9}$F_{1}$                       & \textbf{71.8}                 & 71.7                 & 61.2 $\pm$ 1.9       & \textbf{61.6 $\pm$ 2.8} & 55.5 $\pm$ 5.8       & \textbf{56.5 $\pm$ 5.7} \\\hline
\end{tabular}
\caption{{\color{newtext} Results on additional data \& tasks; finetuning directly from ImageNet pretrained weights (No SSL) vs finetuning after MoCo V2 pretraining. In each experiment, we state the model architecture placed after the ResNet50 backbone, the SSL dataset used to pretrain the backbone, and the task and metric under consideration. For each dataset, we also conduct experiments with 3 subsets of labeled videos used for training.}}

\label{tab:additional_main}
\end{table*}

{\color{changetext}The first is a \textit{saturation} phenomenon, apparent after 10 unlabeled videos; while going from 1 unlabeled video to 10 clearly improves feature quality (phase recognition, finetuning on 5 labeled: +3.1\% $F_{1}$; tool presence detection, finetuning on 5 labeled: +9.3\% mAP), results for 10 and up carry more ambiguity, with large differences depending on the task. While phase recognition performance slows down but still increases by a noticeable amount (e.g. finetuning on 5 labeled, +4.1\% $F_{1}$ from 10 to 80), tool presence detection completely halts.}

The second is \textit{dilution} by labeled data: using larger amounts of annotated videos for finetuning pushes downstream performance closer to its limits, which tends to equalize the effect of adding unannotated videos. {\color{changetext}For example, for phase recognition from 1 to 80 unlabeled videos, $F_{1}$ score increases by 7.2\% with 5 labeled but only by 2.7\% with 40 labeled. }Dilution is much stronger for tool presence detection: from 1 to 80 unlabeled, the total mAP increase with 5 labeled is \textcolor{changetext}{9.5\%}, while no gain is perceivable at all with 40 labeled.

As evidenced by these observations, the performance growth brought by SSL can slow down as the unlabeled data supply increases, depending on the amount of annotated data available as well as the nature of the task. Tool labels are tied to distinct pieces of visual evidence in the image; their influence on the model's final performance is therefore extremely high, compared to unlabeled videos used in self-supervision. In contrast, phase labels tend to accompany more ambiguous visual cues, which would explain why the advantage of using SSL is much more apparent for surgical phase recognition: a model pretrained with 80 unlabeled videos and finetuned on only 5 labeled videos reaches {\color{changetext}60.3\% $F_{1}$}, which is about the same as a model pretrained with 1 unlabeled but finetuned on 10 labeled. Saturation for phase recognition is also much softer than for tool presence detection, suggesting performance can increase even further with more than 80 videos.

{\color{newtext} \subsection{Generalization study} \label{sec:results_generalization}
Using the same recommended hyperparameters established in Section \ref{sec:hyperparameter_study}, we conduct experiments using MoCo v2 on the collection of datasets presented in Section \ref{sec:generalization}. Results are presented in Table \ref{tab:additional_main} demonstrating how SSL representations could be adapted for data from other sources and for other vision-based tasks. 

\textbf{HeiChole Experiments. } In this first experiment series of the generalization study, we utilize the HeiChole Benchmark for surgical workflow analysis. 
Similar to Cholec80, this HeiChole dataset comprises videos for surgical phase recognition and tool presence detection for laparoscopic cholecystectomy. This serves as an ideal benchmark to evaluate how self-supervised representations learned from similar data (same procedure) could be used to boost performance for vision-based tasks on independently sourced datasets with potentially varying surgical workflows, acquisition methods, instrumentation, etc. Indeed, experiments 1 and 2 in Table \ref{tab:additional_main} reveal significant boosts in performance when initializing from models pretrained on Cholec80 (using SSL) at all considered proportions of labeled data. Most notably, using only 2 labeled videos, we observe boosts in performance of $11.4\%$ for phase recognition and $5.2\%$ for tool presence detection. Based on the official leaderboard of the HeiChole challenge, presented in Table \ref{tab:heichole_leaderboard}, this would have positioned our method in $1^{st}$ place for the tool presence detection task and $4^{th}$ for surgical phase recognition using only a simple model architecture. These results strongly exemplify the impact that SSL methods, such as the ones investigated in this article, could have on learning from small datasets and datasets with underrepresented characteristics, problems endemic to surgical data science \citep{sds_2}.

\textbf{CATARACTS Experiments. } Similar to the HeiChole benchmark, the CATARACTS dataset introduces two similar tasks for surgical workflow recognition but with two notable differences: (1) The CATARACTS datasets depict scenes from cataract surgery procedures with a strikingly different appearance and workflow from laparoscopic cholecystectomy (2) The temporal task introduced with this dataset is surgical step recognition, which normally refers to finer temporal segments than surgical phases \citep{mascagni2022computer}. This series of experiments reveals two important findings. Firstly, unlike the HeiChole experiments, models pretrained on Cholec80 (Table \ref{tab:additional_main}, experiments 3 and 5) consistently perform worse than models initialized from Imagenet (``No SSL''). This may be attributed to the significantly distinct and specific visual appearance of Cholec80 scenes serving as a confounding factor when learning representations. However, we do note that when initializing from SSL weights learned on CATARACTS, we see consistent boosts of $\sim1-4\%$ compared to Imagenet initializations across both the downstream tasks. This provides an indication that the SSL setup presented in this work could be adapted to other surgical datasets without further hyperparameter tuning for the pretraining stage.

\textbf{CholecT50 Experiments. } In this series of experiments, we aim to illustrate how self-supervised representations could also help in more difficult workflow tasks like action recognition. To this end, we evaluate performance on CholecT50, a large dataset of surgical actions annotated on videos sourced from the same hospital as Cholec80. Note that the action triplet recognition task on CholecT50 is performed twice (Table \ref{tab:additional_main}, experiments 7 and 8): once using a simple linear head, then a second time with \cite{rdv}'s Rendezvous (RDV) head. In both settings, we observe consistent and marked boosts in performance at all proportions of labeled data demonstrating the utility of these methods across model design choices. Most impressively, utilizing a previously published architecture \citep{rdv} with a generic initialization of features would have placed $3^{rd}$ (Table \ref{tab:cholectriplet_leaderboard}) in the CholecTriplet 2021 challenge \citep{nwoye2022cholectriplet2021}, further illustrating the value that SSL could bring to the surgical data science community.

\textbf{Segmentation Experiments. } Here, we aim to explore how self-supervised representations also have utility for tasks requiring more spatial reasoning than frame-level classification. To this end, we use two surgical semantic segmentation datasets: Endoscapes, consisting of laparoscopic cholecystectomy videos sourced from the same hospital as Cholec80, CaDIS 8 classes and CaDIS 25 classes, containing cataract surgery videos. Consistently, across all three segmentation tasks and labeled data settings, we observe trends consistent with previous findings:  pretraining models using SSL deliver boosts in performance. However, the performance boosts are generally less pronounced than the other considered image recognition tasks. This may be because the considered SSL methods define the learning problem by considering global-level features from the complete image. However, semantic segmentation requires more dense spatial reasoning. More specific architectures choices \citep{caron2021emerging} or SSL methods \citep{wang2021dense, xie2022simmim} could further improve downstream segmentation performance.


}

\begin{table}[]
  \centering
  \begin{tabular}{lr}
  \hline
  \multicolumn{2}{c}{\textbf{CholecTriplet 2021 challenge leaderboard}} \\ \hline
  Rank & Triplet recognition mAP \\ \hline
  $1^{st}$    & 38.1                        \\
  $2^{nd}$    & 35.8                        \\
  \rowcolor{pink} \textbf{MoCo V2 - RDV head}    &  \textbf{35.7}           \\
  $3^{rd}$    &  32.9                       \\
  $4^{th}$ \textit{(RDV baseline)} & 32.7                \\ \hline
  \end{tabular}
  \caption{{\color{newtext} Comparison of MoCo v2 pretraining against the official top 4 entries in the 2021 CholecTriplet challenge.}}

\label{tab:cholectriplet_leaderboard}
\end{table}

\begin{table}[]
  \centering
  \begin{tabular}{lr|lr}
  \hline
  \multicolumn{4}{c}{\textbf{HeiChole Benchmark}} \\ \hline
  Rank & Phase ($F_{1}$) & Rank & Tool ($F_{1}$) \\ \hline
  1                                 & 68.8                            & \cellcolor{pink} \textbf{MoCo V2} & \cellcolor{pink} \textbf{66.9} \\
  2                                 & 65.4                            & 1                                 & 63.8 \\
  3                                 & 65.0                            & 2                                 & 63.0 \\
  \cellcolor{pink} \textbf{MoCo V2} & \cellcolor{pink} \textbf{64.7}  & 3                                 & 58.2 \\
  4                                 & 63.6                            & 4                                 & 50.1 \\ \hline
  \end{tabular}
  \caption{{\color{newtext} Comparison of MoCo v2 pretraining against the official top 4 entries for the phase and tool tasks in the HeiCholec Benchmark (EndoVis challenge 2019).}}

\label{tab:heichole_leaderboard}
\end{table}

\section{Conclusion}
Despite major progress in the field of self-supervised representation learning over the last several years, its adoption into label-scarce fields like surgery, where it could perhaps have the most significant impact, has been slow. This could be due to the demonstrably heavy reliance on hyperparameter choices that SSL methods demand. In this paper, we conduct an extensive benchmark study to methodically identify effective hyperparameter settings for the task of surgical phase recognition and tool presence detection on the Cholec80 dataset. {\color{changetext} From this strong foundation, we deployed SSL on a highly diverse array of surgical datasets, obtaining solid results that support its use for many surgical vision tasks. }

Requiring over 7000 GPU hours, the hyperparameter study demonstrates that this exploration is pivotal to the practical utility of SSL in settings such as semi-supervised learning. For example, initializing the base architecture using Imagenet weights before SSL pretraining critically provided consistent, marked boosts in performance over all other initializations. While random initialization before performing self-supervised representation learning is the standard practice in other large studies, perhaps because of the relative size of the considered datasets, this example highlights the need for principled, adaptable methods to identify optimal settings for other domains. Additionally, domain characteristics could indicate the most significant parameters to prioritize for searches. For instance, in our experiments, relatively slow motion patterns may explain why sampling frames at higher rates for representation learning provides little to no improvement beyond a certain point.

In the data supply study, SSL pretraining shows promising boosts in performance for all methods, particularly in label-scarce scenarios for both phase recognition and tool presence detection. Interestingly, these methods even outperform state-of-the-art methods for semi-supervised phase recognition using only generic representational features. These results are strongly indicative of the value of targeting surgical applications using these SSL methods, which, within certain limits, can be enhanced by simply incorporating additional unannotated data.

{\color{newtext} The generalization study displays the full strength of SSL, with strong results across many surgical contexts; again with generic features obtained without labels. Excellent robustness is demonstrated when switching to a different clinical center or to another task - even the most fine-grained. Results obtained on cataract surgery with hyperparameters conserved from cholecystectomy are highly encouraging for even more radical generalizations of SSL. Further, experimental validation on public challenges, a popular format to introduce and benchmark new datasets, revealed that even simple model architectures with ``generic'' SSL-based initializations achieve more than competitive results compared to significantly more sophisticated design choices. This is despite a recent survey \citep{challengessurvey} concluding that a median of 80 working hours and 267 GPU hours were dedicated in such challenges to model development and training, respectively.
Overall, this section of the study presents a strong exemplification of the value and impact that SSL methods, such as the ones described in this work, could have on supporting ongoing efforts in surgical data science, where small datasets with underrepresented characteristics and expensive annotations are a common occurrence.}

{\color{changetext} Out of the many possibilities opened up by this study, two stand out as highly promising directions for future work: the first one is federated learning \citep{mcmahan2017communication}, where SSL can play a major role by learning robust features from data scattered across multiple clinical centers \citep{kassem2022federated}.} Another natural progression from this work is to apply these findings to recent work in spatio-temporal representation learning and adapt them to the unique characteristics of surgical videos.

Finally, we note that only a select subset of trends were presented for analysis in this work due to many being results aggregated across methods, splits, or other experimental settings for brevity. With around 500 experiments run over 9000 GPU hours, we will disclose complete results for the experiments conducted in this work, in order to facilitate future research on SSL in surgery. The code, along with results and checkpoints, is available at \url{https://github.com/CAMMA-public/SelfSupSurg}.
\\\\
{\bfseries Acknowledgements} This work was partially supported by French state funds managed by the ANR under references ANR-20-CHIA-0029-01 (National AI Chair AI4ORSafety), ANR-10-IAHU-02 (IHU Strasbourg) and ANR-16-CE33-0009 (DeepSurg). This work has received funding from the European Union's Horizon 2020 research and innovation programme under the Marie Sklodowska-Curie grant agreement No 813782 - project ATLAS. This work was supported by a Ph.D. fellowship from Intuitive Surgical. It was granted access to the HPC resources of IDRIS under the allocations 2021-AD011011638R1, \textcolor{newtext}{2021-AD011011638R2,} 2021-AD011012715, 2021-AD011012832, 2021-AD011011507R1, and 2021-AD011011640R1. \textcolor{newtext}{For evaluation on the HeiChole dataset, we thank Dr. Sebastian Bodenstedt for the timely support.}\\\\
\bibliographystyle{model2-names.bst}\biboptions{authoryear}
\bibliography{sssl}


\clearpage
\newpage
\renewcommand{\thesubsection}{\roman{subsection}}
\appendix
\clearpage
\onecolumn
\setcounter{page}{1}

\section*{\centerline{\large Dissecting Self-Supervised Learning Methods for Surgical Computer Vision}}
\subsection*{\centerline{===== ~~~Supplementary Material~~~ =====}}
\begin{center}
    \rule{\linewidth}{0.88pt}
\end{center}

\section{Implementation details}
\subsection{Training: Self-supervised pretraining}
We use ResNet-50 \citep{He2016} as the base encoder architecture and 3-layer multilayer perceptron (MLP) as the projection head for all the Self-supervised methods. The projection head uses $2$ hidden layers ($2048$, $256$) including ReLU activation and batch normalization with an input layer dimension of size $2048$ and output layer dimension of size $4096$. Specific implementation details for each method are as follows: \textit{MoCo v2} uses a queue of size $65536$ with a decay parameter ($\lambda$) of $0.999$ and temperature ($\tau$) of $0.2$; \textit{SimCLR} uses the temperature ($\tau$) of $0.1$; \textit{SwAV} uses $3$ Sinkhorm-Knopp iterations and regularization parameter of $0.05$; \emph{DINO} uses decay parameter ($\lambda$) of $0.996$, warm-up iterations $7500$, and centering decay parameter $\lambda_c$ of $0.9$.

We conduct all the self-supervised training experiments on four V100 GPUs using SGD optimizer with LARC \citep{you2017large} (``trust'' coefficient $\eta$ = $0.001$), a base learning rate of $0.1$, weight decay of $0.000001$, and momentum of $0.9$. The batch size is set to $64$/GPU ($256$ total batch size) except in the \textbf{Batch size} hyperparameter study where we use batches of size $128$ ($32$/GPU), $256$ ($64$/GPU), $512$ ($128$/GPU) and $1024$ ($256$/GPU). We use the VISSL framework\footnote{\url{https://github.com/facebookresearch/vissl}} to run all the experiments using synchronized batch normalization \citep{peng2018megdet} and automatic mixed precision (AMP) \citep{micikevicius2017mixed}. 

For the \textbf{Initialization} hyperparameter study, we initialize ResNet-50 weights as follows: 1) fully-supervised Imagenet weights (``FS'') from torchvision\footnote{\url{https://download.pytorch.org/models/resnet50-0676ba61.pth}}, 2) self-supervised Imagenet weights (``SS'') for SimCLR from VISSL\footnote{\url{https://dl.fbaipublicfiles.com/vissl/model_zoo/simclr_rn50_800ep_simclr_8node_resnet_16_07_20.7e8feed1/model_final_checkpoint_phase799.torch}}, for MoCo v2 from their Github repo\footnote{\url{https://dl.fbaipublicfiles.com/moco/moco_checkpoints/moco_v2_800ep/moco_v2_800ep_pretrain.pth.tar}}, for SwAV from VISSL\footnote{\url{https://dl.fbaipublicfiles.com/vissl/model_zoo/swav_in1k_rn50_800ep_swav_8node_resnet_27_07_20.a0a6b676/model_final_checkpoint_phase799.torch}}, and for DINO from their Github repo\footnote{\url{https://dl.fbaipublicfiles.com/dino/dino_resnet50_pretrain/dino_resnet50_pretrain.pth}}.

\subsection{Training: Downstream tasks}

\textit{Hyperparameter study}: in all the experiments (Section. \ref{sec:hparam_study}), after the self-supervised training, the trunk of the encoder is frozen and the SSL head is replaced by a linear head that maps the feature vector (2048) to the number of outputs for the relevant task. The linear layer is trained on 1 GPU using an SGD optimizer with a base learning rate of $3e-3$ and $1e-1$ for phase recognition and tool presence detection, respectively. The layer is trained for 30 epochs for phase with a step-wise learning rate decay of 0.3 (milestone: 15 epoch) and 50 epochs for tool with a step-wise learning rate decay of 0.1 (milestones: 20, 30 and 40 epoch). Frames are sampled at 1 fps for the linear evaluation described above.

Phase recognition is formulated as a multi-class classification problem where a weighted cross-entropy loss is minimized. The class weights for phases is computed using median frequency balancing \citep{DepthFergusMFB} on the training set.

Since tool presence detection is a multi-label classification problem, we employ weighted binary cross-entropy loss:

\begin{equation}
 L = \sum^{C}_{c=1}\frac{-1}{N} (W_c y_c \log(\sigma(\hat{y}_c)) + (1-y_c)\log(\sigma(\hat{y}_c))),
 \end{equation}
where $y_c$ is the ground truth tool presence label for class c, $\hat{y}_c$ is the predicted probability for class c, and $W_c$ is the class weight.
\\

\textit{Data supply study}: in the data supply study (Section \ref{sec:scalability_study}), after the SSL training, we follow a similar downstream training setup to the one mentioned above, where we replace the SSL head with a linear layer. However, we train the full model for downstream tasks without freezing the encoder's trunk. For the task of phase recognition, all the experiments are trained on 4 GPUs using SGD optimizer with LARC \citep{you2017large} for 30 epochs. Further, we use augmentations and train the model using a cross-entropy loss. When training the temporal model, the TCN is trained on the features extracted after the phase finetuning using an Adam optimizer \citep{kingma2014adam} for 100 epochs with a base learning rate of $3e-3$ and a learning rate decay of $3e-1$ (milestone: 75 epoch). Frames are sampled at 1 fps for the linear finetuning and TCN training described above.

Experiments involving tool presence detection, the models are trained on 4 GPUs using an Adam optimizer for 50 epochs with a base learning rate of $1e-5$ and learning rate decay of 0.1 (milestone: 25 epoch). As specified above, we use weighted binary cross-entropy loss with the class weights computed using inverse frequency balancing for different percentages and samples of the training data used.

All the above training setting is employed when training models with 100\% of the labeled data. In lower labeled data settings, the number of epochs and milestones are scaled inversely in proportion to the percentage of training data. This provides the models with a similar number of updates irrespective of the amount of labeled data available.
\\

{\color{newtext} \textit{Generalization study}: In this step of our study, presented in Section \ref{sec:generalization} and \ref{sec:results_generalization}, we train all the models for the downstream tasks of phase recognition and tool presence detection (HeiChole and CATARACTS) with the same setup described previously in the `Data supply study' paragraph. To comply with the evaluation process of the HeiChole challenge, the TCN, with the fixed hyperparameters, is trained on features extracted at `native' fps of the challenge. 

We follow the data pipeline introduced in \cite{rdv} where we resize the frame to $448 \times 256$ resolution and apply horizontal flipping and brightness/contrast shift as data augmentation strategies. Hyperparameters for finetuning setup involving RDV head and linear head utilize SGD optimizer with a batch size of $32$ and weight decay $1e-6$. The learning rate for the backbone is kept at $1e-4$ whereas the RDV head and linear head use $1e-2$ as the learning rate. For RDV head we utilize a mixture of linear and exponential learning rate schedulers and train for $100$ epochs with early stopping, whereas for linear head setup, we use a multi-step learning rate scheduler with training epochs set to $40$. All the experiments are run using a single NVIDIA A100 GPU.

On both the Endoscapes and CaDIS datasets, we train the downstream semantic segmentation models with an additional DeepLabv3+ head on top of the encoder (ResNet50). In all these experiments, the model is trained on images resized to $480 \times 270$ for 25 epochs with a batch size of 32 and a learning rate of $3e-4$ on a single NVIDIA V100 GPU. }

\section{Augmentation details}
Table \ref{tab:augs:details} shows the details of all the augmentations used for the self-supervised pretraining and supervised finetuning experiments. 

\begin{table}[ht!]
  \centering
  \small
  \caption{Different image augmentations used in the self-supervised pretraining and supervised finetuning experiments. The Source ``RA'' uses the RandAugment \citep{cubuk2020randaugment} implementations that randomly selects two augmentations from the list and applies the augmentation with a given probability ($prob$), magnitude ($M$), and standard deviation on the magnitude ($M_\sigma$). The source ``torchvision'' uses the torchvision implementation for Random-erasing, Horizontal-flip, and Multi-crop augmentations.}
  \label{tab:augs:details}
  \begin{tabular}{llcc}\toprule
                                           & \textbf{Name}     & \textbf{Source} & \textbf{Parameters}                          \\ \midrule
    \multirow{7}{*}{\textbf{Color}}        & Sharpness         & RA              & $prob=0.5, M=8.0, M_\sigma=0.5$              \\
                                           & Brightness        & RA              & $prob=0.5, M=8.0, M_\sigma=0.5$              \\
                                           & Contrast          & RA              & $prob=0.5, M=8.0, M_\sigma=0.5$              \\
                                           & Color             & RA              & $prob=0.5, M=8.0, M_\sigma=0.5$              \\
                                           & Auto-contrast     & RA              & $prob=0.5, M=8.0, M_\sigma=0.5$              \\
                                           & Equalize          & RA              & $prob=0.5, M=8.0, M_\sigma=0.5$              \\
                                           & Random-erasing    & torchvision     & $prob=0.8, scale=[0.02, 0.1]$                \\\midrule
    \multirow{6}{*}{\textbf{Geometric}}    & Rotate            & RA              & $prob=0.5, M=8.0, M_\sigma=0.5$              \\
                                           & Translate-x       & RA              & $prob=0.5, M=8.0, M_\sigma=0.5$              \\
                                           & Translate-y       & RA              & $prob=0.5, M=8.0, M_\sigma=0.5$              \\
                                           & Shear-x           & RA              & $prob=0.5, M=8.0, M_\sigma=0.5$              \\
                                           & Shear-y           & RA              & $prob=0.5, M=8.0, M_\sigma=0.5$              \\
                                           & Horizontal-flip   & torchvision     & $prob=0.5$                                   \\\midrule
    \multirow{3}{*}{\textbf{Strong Color}} & Posterize         & RA              & $prob=0.5, M=8.0, M_\sigma=0.5$              \\
                                           & Solarize          & RA              & $prob=0.5, M=8.0, M_\sigma=0.5$              \\
                                           & invertion         & RA              & $prob=0.5, M=8.0, M_\sigma=0.5$              \\\midrule
    \multirow{3}{*}{\textbf{Multi Crop}}   & MC2: multi-crop 2 & torchvision     & ${224x224=2}$, $scale = [0.5, 1]$            \\
                                           & MC2: multi-crop 4 & torchvision     & $(224x224=2), (96x96=2)$, $scale = [0.5, 1]$ \\
                                           & MC2: multi-crop 8 & torchvision     & $(224x224=2), (96x96=6)$, $scale = [0.5, 1]$ \\
    \bottomrule
  \end{tabular}
\end{table}

\end{document}